\documentclass[10pt,twocolumn,letterpaper]{article}

\usepackage{titling}

\usepackage{iccv}
\usepackage{times}
\usepackage{epsfig}
\usepackage{graphicx}
\usepackage{bm}
\usepackage{amsmath}
\usepackage{amssymb}
\usepackage{comment}
\usepackage{xcolor}
\usepackage{booktabs,makecell,multirow}
\usepackage{setspace}
\usepackage{threeparttable}
\usepackage{colortbl}
\usepackage{makecell}
\usepackage{wasysym}
\usepackage{authblk}
\usepackage[ruled,linesnumbered]{algorithm2e}
\usepackage[T1]{fontenc}
\definecolor{dark-gray}{gray}{0.40}
\definecolor{light-gray}{gray}{0.9}

\usepackage[breaklinks=true,bookmarks=false]{hyperref}

\iccvfinalcopy 

\def\iccvPaperID{10507} 

\ificcvfinal\fi

\begin{document}

\makeatletter
\renewcommand\AB@affilsepx{\ \ \  \protect\Affilfont}
\makeatother

\title{Taxonomy Adaptive Cross-Domain Adaptation in Medical Imaging via Optimization Trajectory Distillation
	\vspace{-5mm}}

\author[1]{Jianan Fan}
\author[1]{Dongnan Liu}
\author[2]{Hang Chang}
\author[3]{Heng Huang}
\author[4]{Mei Chen}
\author[1]{Weidong Cai\vspace{-2mm}}
\affil[1]{\normalsize University of Sydney} 
\affil[2]{\normalsize Lawrence Berkeley National Laboratory}
\affil[3]{\normalsize University of Maryland at College Park}
\affil[4]{\normalsize Microsoft}
\affil[ ]{\small \texttt{jfan6480@uni.sydney.edu.au\qquad dongnan.liu@sydney.edu.au\qquad  hchang@lbl.gov\qquad henghuanghh@gmail.com\qquad Mei.Chen@microsoft.com\qquad tom.cai@sydney.edu.au}}

\maketitle
\ificcvfinal\thispagestyle{empty}\fi

\begin{abstract}
	The success of automated medical image analysis depends on large-scale and expert-annotated training sets. 
	Unsupervised domain adaptation\;(UDA) has been raised as a promising approach to alleviate the burden of labeled data collection.
	However, they generally operate under the closed-set adaptation setting assuming an identical label set between the source and target domains, which is over-restrictive in clinical practice where new classes commonly exist across datasets due to taxonomic inconsistency.
	While several methods have been presented to tackle both domain shifts and incoherent label sets, none of them take into account the common characteristics of the two issues and consider the learning dynamics along network training.
	In this work, we propose optimization trajectory distillation, a unified approach to address the two technical challenges from a new perspective.
	It exploits the low-rank nature of gradient space and devises a dual-stream distillation algorithm to regularize the learning dynamics of insufficiently annotated domain and classes with the external guidance obtained from reliable sources.
	Our approach resolves the issue of inadequate navigation along network optimization, which is the major obstacle in the taxonomy adaptive cross-domain adaptation scenario.
	We evaluate the proposed method extensively on several tasks towards various endpoints with clinical \textcolor{black}{and open-world} significance.
	The results demonstrate its effectiveness and improvements over previous methods.
	Code is available at https://github.com/camwew/TADA-MI.
\end{abstract}

\section{Introduction}
\label{sec:introduction}

\begin{figure}[!t]
	\centerline{\includegraphics[width=\columnwidth]{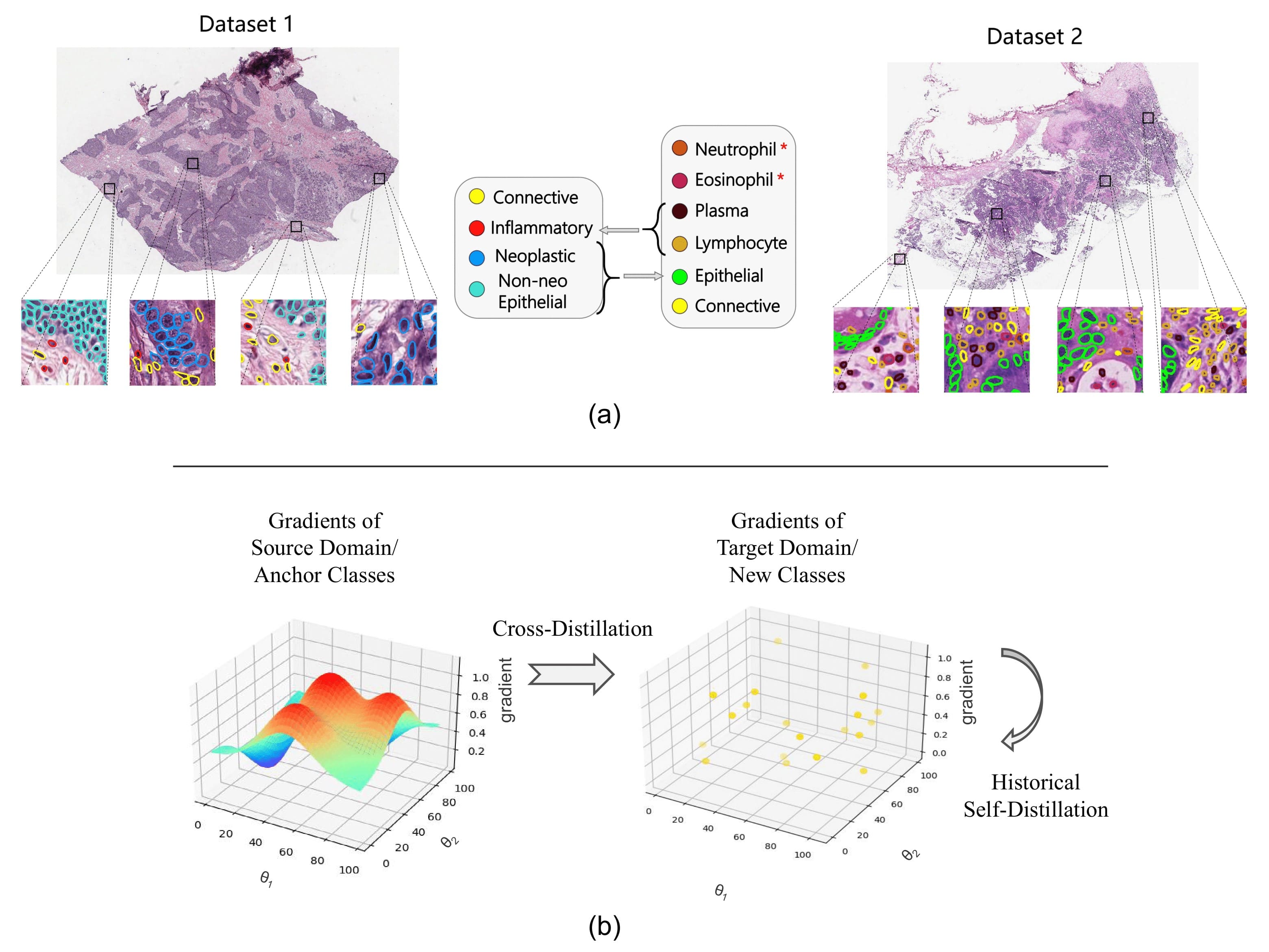}}
	\caption{(a)\;Illustration of the issue of taxonomic inconsistency with nuclei recognition.  The category label sets are incoherent across datasets. Different colours indicate the class each nucleus belongs to. The red asterisks denote novel classes only existing in Dataset 2. 
		(b)\;Concept of the proposed method. We perform optimization trajectory distillation 
		to deliver external navigation for learning target domain and new classes where the optimization steps tend to be restricted and unreliable.}
	\label{fig:problem}
\end{figure}
Automated and objective analysis of medical images is an important research topic and has been explored in various clinical applications \cite{winkler2019association, dominguez2022cross}.
To relieve the burden of acquiring massive annotated data for model development on new domains, several unsupervised domain adaptation\;(UDA) methods \cite{chen2019synergistic, liu2020unsupervised, liu2022decompose} are proposed to mitigate the data distribution bias \cite{torralba2011unbiased} between a richly labeled source domain and a target domain with no explicit supervision.

However, these works assume a closed-set adaptation setting that the source and target domains should necessarily share the identical category label space and definition \cite{panareda2017open}.
The restriction limits their practical applicability in \emph{the clinical wild} \cite{gamper2019pannuke}.
Unlike natural images where the definitions of different entity categories\;(\emph{e.g.}, cat v.s.\;dog) are structured and globally unified, in the medical domain, inconsistency in taxonomy across different countries or institutes is a common issue compromising the feasibility of cross-domain adaptation \cite{dodd2018taxonomy}.
Take nuclei recognition in histology images as an example, the ambiguous biological characteristics of cells and distinct clinical usages of the analysis outcomes result in a lack of a unified categorization schema \cite{gamper2019pannuke, graham2021lizard}.
With specific clinical purposes and down-stream applications, different datasets could categorize nuclei with disparate granularities of biological concepts.
Besides, as medical research evolves rapidly, novel cell and disease classes could be discovered and form datasets with continually expanding category label sets \cite{lupton2012medicine}.
This motivates us to develop a generalized adaptation method with the capability to tackle both data distribution bias and category gap for more flexible UDA in clinical practice.

To this end, 
we study the problem of taxonomy adaptive domain adaptation \cite{gong2022tacs}.
It assumes the target domain could adopt a label space different from the source domain.
Following UDA, a labeled source dataset and unlabeled samples from the target domain are available during training.
Additionally, to recognize the novel/fine-grained categories which are non-existent or unspecified in the source domain, a few samples of those target-private categories are annotated and utilized as exemplars.
The technical challenge for this setting lies in how to alleviate domain shifts and concurrently learn new classes with very limited supervision.
%
Recently, several methods are proposed towards a similar goal \cite{kundu2020class, guo2020broader}.
However, they typically address the two issues individually in separate contexts and fail to design a unified paradigm according to their common characteristics \cite{gong2022tacs}, which could incur a subtle collision across issues since their objectives are non-relevant \cite{yu2020gradient}.
Besides, existing works either focus on cross-domain/class alignment in the feature space \cite{luo2017label, liang2021boosting} or resort to model output-based regularization such as self-supervision \cite{phoo2021selftraining}, whereas those approaches suffer from the equilibrium challenge of adversarial learning \cite{arora2017generalization} and the low-quality pseudo-labels \cite{wang2020unsupervised}.

In this work, we present a unified framework for taxonomy adaptive cross-domain adaptation from a new perspective, \emph{i.e.}, via optimization trajectory distillation.
Our strategy is motivated by a common challenge existing in both cross-domain and small-data learning regimes, which is the inadequate navigation in the course of network optimization.
For cross-domain learning, the unstable feature alignment procedure and noisy pseudo-labels tend to induce error accumulation along network training \cite{zhang2019category}.
Similarly, with limited support samples in new class learning, the optimization of network is inclined to step towards restricted local minima and cannot cover a globally-generalizable space \cite{wang2020generalizing}.
To this end, we propose to
exploit the optimization trajectory from a reliable ``teacher'' to provide external guidance for regularizing the learning dynamics of insufficiently annotated domain and classes, as illustrated in Fig.\;\ref{fig:problem}(b).

Our method consists of two key components, \emph{i.e.,} cross-domain/class distillation and historical self-distillation.
\textbf{(i)}\;Cross-domain and cross-class distillation aim to leverage the optimization trajectory from the richly labeled domain and classes to serve as the ``teacher''.
Motivated by the Neural Tangent Kernel\;(NTK) theory \cite{jacot2018neural}, we characterize the network optimization trajectory through gradient statistics.
Then, given the observation that the subspace spanned by the gradients from most iterations is generally low-rank \cite{li2020hessian, azam2022recycling}, we design a gradient projection approach to suppress the noisy signals in stochastic gradients and rectify the distillation process.
Thereafter constraints are imposed on the gradient statistics of target domain and new classes to calibrate their training dynamics towards domain-invariant and unbiased learning procedure.
\textbf{(ii)}\;Historical self-distillation further drives the optimization paths of model to converge towards flat minima.
It is found that the flatness of loss landscapes is strongly related to the model's robustness and generalizability \cite{izmailov2018averaging},
while how to take advantage of the insight to tackle domain shifts and limited supervision is under-explored.
We propose to exploit the historical gradients to construct the informative low-rank subspaces and then perform gradient projection to alleviate loss sharpness.
It compensates for the intense and out-of-order optimization updates incurred by inadequate regularization and leads to better generalization.

Our prime contributions are as follows:
(1)\;We introduce a more generalized cross-domain adaptation paradigm for medical image analysis in which both data distribution bias and category gap exist across the source and target domains.
(2)\;We leverage insights from recent learning theory research and propose a novel dual-stream optimization trajectory distillation method to provide external navigation in network training.
We perform theoretical justifications from two perspectives to illustrate the merits of our method.
(3)\;Experiments on various benchmarks validate the effectiveness and robustness of our proposed method and its improvements over existing approaches.

\section{Related Works}

\noindent\textbf{Domain Adaptation} \ 
Domain adaptation\;(DA) aims to mitigate the data distribution bias \cite{ganin2016domain}.
The classic semi-supervised/unsupervised DA focuses on the scenario where the target domain has an identical label space to the source domain \cite{saito2019semi, du2021cross, li2022domain}.
The over-restricted precondition cannot suffice the demand of clinical usages \cite{xu2021class}.
There exist several efforts to take a step further than the closed-set adaptation, such as open-set DA \cite{panareda2017open} and universal DA \cite{you2019universal}.
However, the aim of those works is solely to detect the previously unseen classes, instead of learning to separately identify each new class with few supports.
Some recent studies suggest to explicitly recognize the target-private categories.
\cite{kundu2020class} performs feature projection and alignment based on the prototypical networks \cite{snell2017prototypical}.
\cite{gong2022tacs} combines pseudo-labeling and contrastive learning to combat domain bias and label gap.
Nevertheless, those works are limited to tackling the two issues individually, which fails to exploit their common characteristics and propose a unified solution.

\noindent\textbf{Cross-Domain Few-Shot Learning} \ 
The technical challenge we need to resolve is similar to an emerging research topic, \emph{i.e.,} cross-domain few-shot learning \cite{guo2020broader}.
The mainstream methods for this challenge include self-training \cite{phoo2021selftraining, islam2021dynamic, yuan2022task}, contrastive learning \cite{das2022confess, gong2022tacs}, feature alignment/augmentation \cite{Tseng2020Cross-Domain, chen2022cross, hu2022adversarial}, and channel transformation \cite{li2022crossfew, luo2022channel}.
Those methods either focus on feature-level modulation or turn to output-level self-supervision and fine-tuning, which could bring instability along optimization \cite{arora2017generalization, wang2020unsupervised}.
In this work, we propose a novel gradient-based framework to transfer knowledge in both cross-domain and few-shot settings.
It implicitly characterizes the information in both feature space and output space.

\noindent\textbf{Gradient Manipulation}
In current deep learning systems where gradient descent serves as the most popular training algorithm, the directions and amplitudes of network optimization steps are embedded in the gradient space \cite{chen2018gradnorm}.
Gradient manipulation refers to directly modulating the optimization gradients for improved learning process \cite{mansilla2021domain}.
Recently, it is introduced to mitigate the data distribution shift and shows promises in UDA and domain generalization \cite{du2021cross, gao2021gradient, shi2022gradient}.
However, those works are still restricted by the closed-set condition and do not consider the low-rank nature of gradients \cite{li2020hessian}, which make their approaches vulnerable to noises in the small-data learning regime.

\noindent\textbf{Flatness in Optimization} \ 
In previous machine learning research, the connection between convergence to flat minima in optimization and model generalizability has long been established \cite{hochreiter1997flat, keskar2017on}.
Theoretical and empirical studies prove that the flatness of loss landscapes highly correlates with generalization capability \cite{caldarola2022improving}.
\cite{cha2021swad} averages the network weights of multiple checkpoints to prompt generalization across domains.
Different from their post-hoc averaging approach operating on fixed models, we propose to regularize the intermediate dynamics of network optimization to achieve flat minima.

\section{Method}
\subsection{Problem Statement}
The taxonomy adaptive cross-domain adaptation problem can be formalized as follows.
There exists a labeled source dataset $\mathcal{D}_s=\{(\mathbf{x}_s, \mathbf{y}_s)\}$, an unlabeled target dataset $\mathcal{D}_t^u=\{(\mathbf{x}_t^u)\}$, and a few-shot labeled support set from the target domain $\mathcal{D}_t^l=\{(\mathbf{x}_t^l, \mathbf{y}_t^l)\}$.
Samples from source and target datasets are drawn from different data distributions, \emph{i.e.}, $\mathbf{x}_s\sim\mathcal{P}_s$, $\mathbf{x}_t\sim\mathcal{P}_t$, $\mathcal{P}_s\neq \mathcal{P}_t$.
The label sets of source and target datasets are denoted as $\mathcal{C}_s$ and $\mathcal{C}_t$.
Due to the novel/fine-grained categories only existing in the target dataset, we have $\mathcal{C}_s \neq \mathcal{C}_t$.
The shared and target-private new classes are indicated as $\overline{\mathcal{C}} = \mathcal{C}_s\cap\mathcal{C}_t$ and $\hat{\mathcal{C}} = \mathcal{C}_t \backslash \overline{\mathcal{C}}$.
We call $\overline{\mathcal{C}}$ existing in both datasets as anchor classes since they can be used as anchors to facilitate knowledge transfer.
The goal is to train a model jointly on $\mathcal{D}_s, \mathcal{D}_t^u, \mathcal{D}_t^l$ which performs inferences on data from $\mathcal{P}_t$ with labels in $\mathcal{C}_t$.

\begin{figure*}[!t]
	\centerline{\includegraphics[width=2\columnwidth]{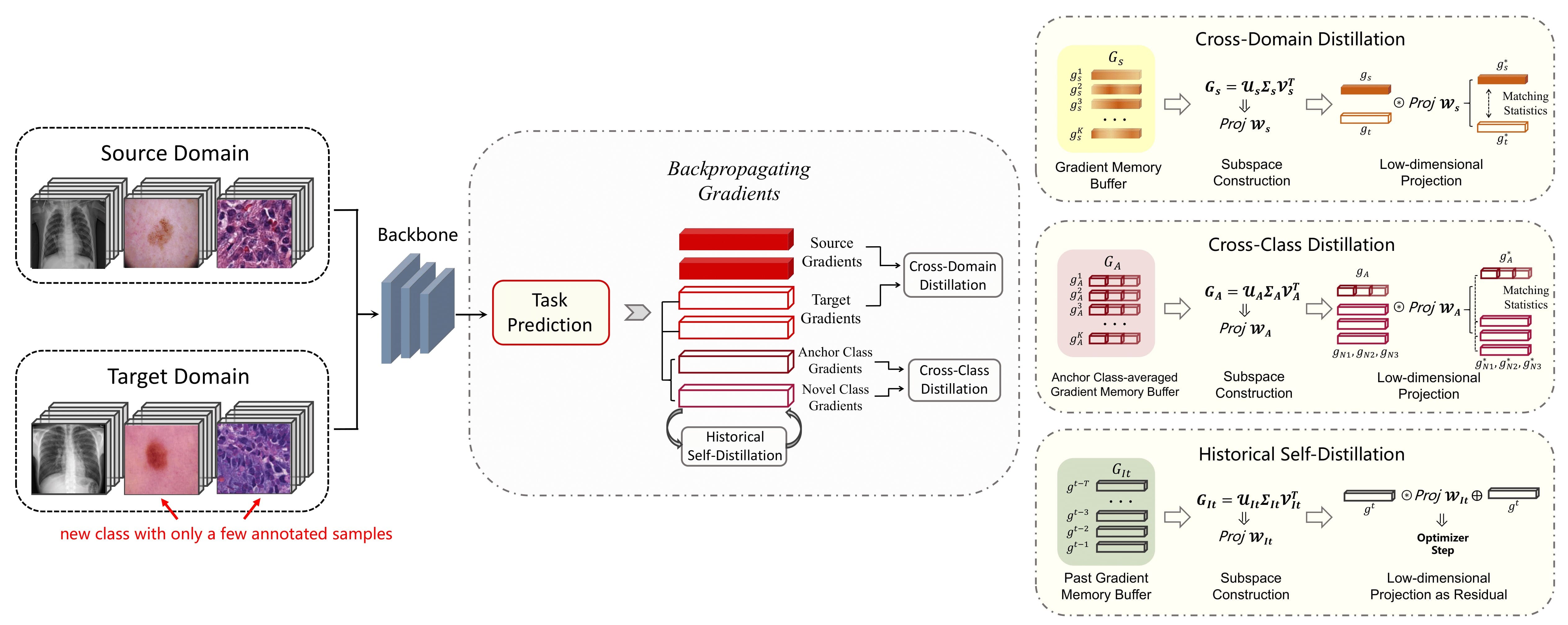}}
	\caption{\textcolor{black}{Overview of our proposed method. }
		We devise the cross-domain and cross-class distillation module as well as the historical self-distillation module to transfer the model optimization knowledge from reliable sources by imposing constraints on the gradient descent steps \emph{w.r.t.} the insufficiently supervised target domain and new classes.}
	\label{fig:model_main}
\end{figure*}

\subsection{Characterization of Optimization Trajectory}
\label{sec:waytopresentgradient}
The technical challenge for the identified problem lies in overcoming both domain shifts and overfitting incurred by scarce training samples.
As discussed in Section \ref{sec:introduction}, those issues could lead to unreliable navigation during neural network optimization \cite{Ge2020Mutual, cheraghian2021synthesized}.
Therefore, we propose to distill the optimization trajectory knowledge from well-annotated sources to regularize the training dynamics of target domain and new classes.

Recently, NTK theory \cite{jacot2018neural} has been regarded as a tool to dictate the evolution of neural networks.
For a network $f$ parameterized by $\theta$, its training dynamics can be described by a kernel $K$.
Given a training set with $N$ samples, the kernel can be defined as an $N\times N$ matrix with $K(i, j; \theta) = \nabla_\theta f(x^i; \theta)^\top \cdot \nabla_\theta f(x^j; \theta)$, where $x^i$ and $x^j$ are two input samples.
It indicates the strong correlations between the network's optimization trajectory and the gradients \emph{w.r.t.} its parameters.
We thus propose to collect a set of intermediate gradients during training and capture their statistics to characterize the optimization trajectory.
Specifically, the second-order Taylor expansion shows that:
\begin{equation}\begin{split}\label{equ:taylor}
		f(x; \theta) = &f(x; \theta^*) + (\theta- \theta^*)^\top\cdot\nabla_\theta f(x; \theta^*) \cr +& \frac{1}{2}(\theta- \theta^*)^\top\cdot \nabla_\theta^2  f(x; \theta^*)\cdot(\theta- \theta^*) \cr +& \mathcal{O}(\left \|\theta- \theta^*\right\|^2),
\end{split}\end{equation}
which motivates us to model the first- and second-order derivatives\;(Hessian) of $f$ to represent its learning dynamics.
We use the mean of gradients to indicate the first-order derivative and follow \cite{jastrzkebski2018three} to approximate the second-order derivative with gradient variances.

\subsection{Cross-Domain and Cross-Class Distillation}
\label{sec:cross-distill}
Gradient-based optimization algorithms depend on a large number of update steps supervised by sufficient labeled data \cite{ravi2017optimization}.
In domain-shifted and small-data regimes, those algorithms suffer from noisy update signals and overfitting, without guarantee to converge towards generalizable minima \cite{guo2020broader}.
To this end, we propose to distill the training dynamics from the label-rich source domain and anchor classes to the target domain and new classes for external regularization by leveraging their intermediate gradients.
We firstly compute the gradients $\bm{g}$ for each domain and class by backpropagating the obtained losses.
With a general backbone $\mathbf{F}$ for feature extraction and a task-specific predictor $\mathbf{C}$ for sample-/pixel-wise classification, the individual gradients can be expressed as:
\begin{equation}\label{equ:gradient_comput}
	\bm{g}:=[\frac{\partial\mathcal{L}(\mathbf{C}(\mathbf{F}(x)), y)}{\partial\theta_{\mathbf{C}_1}},
	\cdots,\frac{\partial\mathcal{L}(\mathbf{C}(\mathbf{F}(x)), y)}{\partial\theta_{\mathbf{C}_w}}],
\end{equation}
where $\mathcal{L}$ is the loss function, $w$ indicates the total number of parameters in $\mathbf{C}$, $x$ and $y$ are training data and the corresponding label.
For unlabeled samples $\mathbf{x}_t^u$ in the target dataset, online pseudo-labels $\tilde{\mathbf{y}}_t^u$ are used for loss calculation.
Then, as shown in Fig.\;\ref{fig:model_main}, we aggregate the individual-level gradients in a memory buffer and retrieve the domain- and class-level gradient statistics to model the global first- and second-order derivatives.
According to Section\;\ref{sec:waytopresentgradient}, for gradients \emph{w.r.t.} domain/class $\pi$ collected from a training set with $n_\pi$ samples, we derive their mean and variance as:
\begin{small}\begin{equation}\label{equ:gra_mean}
		\bm{g}_\pi^m=\frac{1}{n_\pi}\sum\nolimits_{i=1}^{n_\pi}\bm{g}_\pi^i,
	\end{equation}
	\begin{equation}\label{equ:gra_var}
		\bm{g}_\pi^v=\frac{1}{n_\pi-1}\sum\nolimits_{i=1}^{n_\pi}(\bm{g}_\pi^i-\bm{g}_\pi^m)^2.
\end{equation}\end{small}%
Those metrics are measured for each domain and class separately.

%
%
However, the gradients obtained from insufficiently annotated domain and classes are dubious and error-prone \cite{wang2020generalizing}.
To alleviate the issue, we take inspiration from recent learning theory findings that stochastic gradients along the optimization steps are generally low-rank \cite{li2020hessian},
which indicates that the optimization process is governed by the top gradient eigenspace.
Specifically, we devise a gradient projection approach to suppress the noisy model update signals in particular to the target domain and new classes by projecting their gradients on the reliable subspaces identified by the source domain and anchor classes with sufficient supervision.
Our approach involves three iterative steps, \emph{i.e.}, subspace construction, gradient projection, and gradient statistics matching.
Firstly, to identify the principal eigenspace, we perform Singular Value Decomposition\;(SVD) on the aggregated gradients from the source domain and anchor classes.
We denote gradients collected from source domain, target domain, anchor class, and new class as $\bm{g}_s$, $\bm{g}_t$, $\bm{g}_A$, and $\bm{g}_N$.
Given the sets of gradients $\bm{G}_s=[\bm{g}_s^1, \bm{g}_s^2, \cdots, \bm{g}_s^K]$ and $\bm{G}_A=[\bm{g}_A^1, \bm{g}_A^2, \cdots, \bm{g}_A^K]$ stored in the memory buffer of volume $K$, we apply SVD:
\begin{small}\begin{equation}\label{equ:svd}
		\bm{G}_s=\bm{U}_s\bm{\Sigma}_s\bm{V}_s^\top,
		\qquad
		\bm{G}_A=\bm{U}_A\bm{\Sigma}_A\bm{V}_A^\top,
\end{equation}\end{small}%
where $\bm{U}$, $\bm{V}$, and $\bm{\Sigma}$ contain left singular vectors $\bm{u}^i$, right singular vector $\bm{v}^i$, and non-negative singular values $\bm{\sigma}^i$, respectively.
To capture the overall characteristics, we use the gradients averaged along all anchor classes to represent $\bm{G}_A$.
Next, we adopt the low-rank matrix approximation to select the top-$r$ significant left singular vectors in $\bm{U}_s$ and $\bm{U}_A$ based on the following criteria:
\begin{small}\begin{equation}\label{equ:topk}
		\left \|(\bm{G}_s)^{r_s}\right\|_F^2\geq\tau\left \|\bm{G}_s\right\|_F^2,
		\quad
		\left \|(\bm{G}_A)^{r_A}\right\|_F^2\geq\tau\left \|\bm{G}_A\right\|_F^2,
\end{equation}\end{small}%
where $(\bm{G})^r=\sum\nolimits_{i=1}^{r}\bm{\sigma}^i\bm{u}^i\bm{v}^{i\top}$, $\left \|\cdot\right\|_F$ denotes the Frobenius norm, $\tau$ is a threshold to ensure most information is preserved and is set to 0.98.
The principal subspace and the corresponding projection matrix are thereby constructed as
$\bm{M}_s=[\bm{u}_s^1,\bm{u}_s^2,\cdots,\bm{u}_s^{r_s}]$ and $\bm{M}_A=[\bm{u}_A^1,\bm{u}_A^2,\cdots,\bm{u}_A^{r_A}]$.
Afterward, all gradients are projected on the identified subspace as shown in Fig.\;\ref{fig:model_main}:
\begin{footnotesize}\begin{equation}\label{equ:project}\begin{split}
			\bm{g}_s^*=\bm{M}_s\bm{M}_s^\top\bm{g}_s,&
			\qquad
			\bm{g}_t^*=\bm{M}_s\bm{M}_s^\top\bm{g}_t,\cr
			\bm{g}_A^*=\bm{M}_A\bm{M}_A^\top\bm{g}_A,&
			\qquad
			\bm{g}_N^*=\bm{M}_A\bm{M}_A^\top\bm{g}_N.
\end{split}\end{equation}\end{footnotesize}%
Then following Eq.(\ref{equ:gra_mean})(\ref{equ:gra_var}), we minimize the discrepancy between the statistics\;(\emph{i.e.}, mean and variance) of projected gradients to distill the learning dynamics from the source domain and anchor classes to the target domain and new classes.
The overall training objective can be formulated as:
\begin{small}\begin{equation}\begin{split}\label{equ:objective}
			&\mathop{\mathrm{min}}\limits_\theta\ \ \mathcal{L}_{ERM}+\lambda\Big\{\left \|(\bm{g}_s^{m*}-\bm{g}_t^{m*})\right\|_2^2 + \left \|(\bm{g}_s^{v*}-\bm{g}_t^{v*})\right\|_2^2 \cr
			&+ \frac{1}{n_{new}}\sum\nolimits_{i=1}^{n_{new}}\big[\left \|(\bm{g}_A^{m*}-\bm{g}_{Ni}^{m*})\right\|_2^2 + \left \|(\bm{g}_A^{v*}-\bm{g}_{Ni}^{v*})\right\|_2^2 \big]\Big\}.
\end{split}\end{equation}\end{small}%
Here $\mathcal{L}_{ERM}$ denotes the empirical risk loss, which is implemented with cross-entropy loss for classification and Dice loss \cite{milletari2016v} for segmentation.
$\lambda$ is a balancing coefficient, $n_{new}$ is the number of new classes in the target domain.
It is noted that there exists discrepancy between the anchor and new classes in semantic concepts, which makes feature alignment across classes harmful due to negative transfer \cite{wang2019characterizing}.
However, aligning class-wise learning dynamics by enforcing the similarity between their gradients could contribute to lower empirical error on new classes.
We prove this property theoretically in the supplementary material.

\subsection{Historical Self-distillation}
Inferior generalization behaviour is an outstanding challenge for taxonomy adaptive cross-domain adaptation \cite{gong2022tacs}.
When domain shifts and unseen classes exist, a trained model could only maintain its performance on test samples sharing multiple similar attributes with the training data but cannot generalize well to the disparate ones \cite{zuo2021challenging}.
Motivated by the connection between flat minima and generalizability \cite{keskar2017on}, we devise the historical self-distillation module to smooth the optimization path towards a well-generalizable solution by calibrating gradient descent steps.

To reach flat loss landscapes \cite{chaudhari2017entropysgd}, assuming a local loss function $\psi$ and a network $f$ parameterized by $\theta$, instead of minimizing the original loss $\psi(f(x; \theta))$ only, the optimization objective should be:
\begin{equation}
	\mathop{\mathrm{min}}\limits_\theta\  -\log\int_{\left\|\theta-\theta^*\right\|\leq\Gamma}\exp(-\psi(f(x; \theta^*))-\gamma\left\|\theta-\theta^*\right\|)\ d\theta^*,
\end{equation}%
where $x$ indicates random input samples, $\Gamma$ is defined as the width of a local parameter valley, $\gamma$ is a balancing weight.
In other words, when $\theta$ and $\theta^*$ are neighbouring, the following criteria is required to be satisfied:
\begin{equation}
	\left\|\nabla_\theta\psi(f(x; \theta))-\nabla_\theta\psi(f(x; \theta^*))\right\|\leq\beta\left\|\theta-\theta^*\right\|,
\end{equation}%
where $\beta$ is a smoothness factor.
To this end, we propose to impose regulations on the optimizer steps by enforcing minima with uniformly-distributed gradients.
Given that the gradients from most iterations are generally low-rank \cite{li2020hessian, azam2022recycling},
we exploit past gradients to identify the low-dimensional subspace which indicates the principal gradient distribution of local minima and then project current gradients on the constructed subspace to exclude sharp and noisy signals.
Specifically, we collect and store individual gradients for each iteration along the training procedure, denoted as $\bm{G}_{It}=[\bm{g}_{It}^{t-T}, \bm{g}_{It}^{t-T+1}, \cdots, \bm{g}_{It}^{t-2}, \bm{g}_{It}^{t-1}]$, where $t$ is the index of current iteration, $T$ is the size of memory buffer.
Thereafter we perform SVD on the set of gradients $\bm{G}_{It}$ to construct the historical subspace.
Similar to Eq.(\ref{equ:svd})(\ref{equ:topk})(\ref{equ:project}), the top-$r$ left singular vectors of the decomposed gradient set are concatenated to formulate the projection matrix $\bm{M}_{It}$.
Then for upcoming gradients obtained from backpropagation in each iteration, we project the gradients on the identified low-rank subspace and consider them as residuals.
The subsequent optimizer step is conducted with the sum of the original and rectified gradients, as shown in Fig.\;\ref{fig:model_main}.
Take vanilla stochastic gradient descent as an example, with mini-batch gradients $\tilde{\bm{g}}$ in each iteration, the parameters $\theta$ of the optimized network are updated by:
\begin{equation}\label{equ:step}
	\theta:=\theta-\frac{1}{\kappa}\cdot\eta\cdot (\bm{M}_{It}\,\bm{M}_{It}^\top+\kappa)\cdot\tilde{\bm{g}},
\end{equation}%
where $\eta$ is the learning rate, $\kappa$ is a trade-off parameter.
As training proceeds, the principal subspace and the corresponding projection matrix are periodically renewed by applying SVD on the recently collected gradients.
The overall training procedure is summarized in the supplementary material.

\begin{table*}[!t]
	\centering
	\fontsize{8.5}{9.5}\selectfont
	\begin{threeparttable}
		\caption{Comparison results of our proposed method against other state-of-the-art methods for nuclei segmentation and recognition. The metrics with asterisk\,(*) are calculated on target-private new classes only. The subscripts denote the standard deviations. The best and second-best results are highlighted in bold and brown, respectively.}
		\label{tab:nuclei_result}
		\setlength{\tabcolsep}{2mm}{
			\begin{tabular}{ccccccccc}
				\toprule[0.5mm]
				\multirow{3}{*}{\footnotesize \textbf{Methods}}
				&\multicolumn{4}{c}{\textbf{5-shot}}
				&\multicolumn{4}{c}{\textbf{10-shot}}\cr
				\cmidrule(lr){2-5} \cmidrule(lr){6-9} 
				&mF1&mF1*&mPQ&mPQ*&mF1&mF1*&mPQ&mPQ*\cr
				\midrule[0.3mm]				
				Sup-only
				&28.08$_{0.89}$&21.46$_{0.36}$&12.35$_{0.53}$&12.23$_{0.45}$
				&30.90$_{0.66}$&25.23$_{0.83}$&14.18$_{0.24}$&14.94$_{0.32}$\cr
				Multi-task\,\cite{sener2018multi}
				&33.85$_{1.26}$&18.15$_{1.08}$&18.58$_{0.42}$&10.77$_{0.35}$
				&35.15$_{0.94}$&21.29$_{0.62}$&19.14$_{0.27}$&12.89$_{0.34}$\cr
				\midrule[0.2mm]			
				DANN\,\cite{ganin2016domain}
				&32.76$_{1.41}$&16.52$_{1.10}$&18.31$_{0.66}$&10.40$_{0.59}$
				&35.76$_{1.35}$&23.12$_{0.88}$&19.70$_{0.60}$&14.14$_{0.42}$\cr
				CGDM\,\cite{du2021cross}
				&35.03$_{1.28}$&21.04$_{0.99}$&19.59$_{0.50}$&13.22$_{0.32}$
				&37.13$_{1.57}$&24.22$_{1.15}$&19.78$_{0.71}$&13.90$_{0.40}$\cr
				\midrule[0.2mm]						
				LETR\,\cite{luo2017label}
				&27.78$_{1.44}$&15.36$_{1.02}$&16.68$_{0.54}$&10.03$_{0.51}$
				&34.85$_{1.54}$&20.03$_{0.73}$&18.18$_{0.59}$&11.79$_{0.27}$\cr
				FT-CIDA\,\cite{kundu2020class}
				&30.51$_{0.92}$&16.55$_{0.80}$&15.97$_{0.39}$&9.42$_{0.40}$
				&32.63$_{1.21}$&21.08$_{0.95}$&17.09$_{0.47}$&11.96$_{0.49}$\cr
				STARTUP\,\cite{phoo2021selftraining}
				&37.03$_{0.58}$&22.44$_{0.76}$&20.40$_{0.33}$&13.82$_{0.30}$
				&\textcolor{brown}{40.85$_{0.50}$}&25.92$_{0.91}$&\textcolor{brown}{23.37$_{0.19}$}&16.64$_{0.24}$\cr	
				DDN\,\cite{islam2021dynamic}
				&\textcolor{brown}{37.92$_{0.86}$}&\textcolor{brown}{23.71$_{1.05}$}&\textcolor{brown}{20.62$_{0.57}$}&\textcolor{brown}{13.87$_{0.43}$}
				&40.16$_{0.72}$&\textcolor{brown}{26.94$_{1.08}$}&22.79$_{0.32}$&\textcolor{brown}{16.82$_{0.50}$}\cr	
				TSA\,\cite{li2022crossfew}
				&34.03$_{1.17}$&19.55$_{1.11}$&18.65$_{0.43}$&12.06$_{0.37}$
				&34.27$_{1.05}$&22.58$_{1.20}$&19.78$_{0.61}$&14.24$_{0.73}$\cr	
				TACS\,\cite{gong2022tacs}
				&36.73$_{1.65}$&22.26$_{1.29}$&18.13$_{0.80}$&12.57$_{0.62}$
				&39.29$_{1.48}$&25.84$_{1.13}$&22.27$_{0.68}$&15.98$_{0.56}$\cr	
				\rowcolor{light-gray} Ours
				&\textbf{40.26}$_{0.90}$&\textbf{27.14}$_{0.97}$&\textbf{21.78}$_{0.49}$&\textbf{14.96}$_{0.22}$
				&\textbf{43.88}$_{0.52}$&\textbf{31.43}$_{0.78}$&\textbf{24.81}$_{0.26}$&\textbf{19.35}$_{0.33}$\cr	
				\bottomrule[0.5mm]
		\end{tabular}}
	\end{threeparttable}
\end{table*}
\subsection{Theoretical Analysis}
In this section, we motivate our proposed method theoretically and provide mathematical insights on its merits.

\noindent\textbf{Joint Characterization of Feature and Output Space} \ 
While our optimization trajectory distillation approach proposes a novel perspective that differs from existing works imposing regularization terms in feature and model-output space \cite{kundu2020class, gong2022tacs}, those information is still captured and implicitly modeled in the gradient-based framework.
We prove this property for the classification and segmentation tasks under two commonly adopted loss functions.
The detailed proof is presented in the supplemental material.
It illustrates the superiority of our method as a unified framework that jointly characterizes the feature and output space as well as the learning dynamics.

\noindent\textbf{Impacts on Generalization Error} \ 
In our method, we propose to minimize the discrepancy between the gradient statistics from different domains and classes in the identified principal subspace.
We hereby prove its effectiveness towards a tighter generalization error bound on the target domain and new classes \cite{ben2006analysis}.
Details can be found in the supplemental material.

\section{Experiments}
\subsection{Datasets and Experiment Settings}
To evaluate the effectiveness of our method and its potential clinical implications, we conduct experiments on five medical image datasets in regard to three important down-stream tasks from different clinical scenarios.
\textcolor{black}{Extended experiments on more diverse tasks, including radiology and fundus analysis, as well as general visual task, are presented in the supplemental material to substantiate the broader applicability of our method.}

\noindent\textbf{Nuclei Segmentation and Recognition} \ 
Technically, this task can be formalized as a simultaneous instance segmentation and classification problem \cite{graham2019hover}.
We evaluate our method by considering PanNuke \cite{gamper2019pannuke} and Lizard \cite{graham2021lizard} as the source and target datasets.
They contain 481 and 291 visual fields cropped from whole-slide images with 189,744 and 495,179 annotated nuclei, respectively.
There are six types of nuclei in Lizard and only two of them overlap with the five classes in PanNuke, which suggests four new classes in the target domain.
We employ the widely used Hover-Net \cite{graham2019hover} architecture with a standard ResNet-50 backbone as the base model.
For quantitative evaluation, we follow \cite{graham2021lizard} and report the average F1 and PQ\;(panoptic quality) scores for all classes\;(denoted as mF1 and mPQ) to provide a comprehensive assessment measuring the accuracy of nuclei detection, segmentation, and classification.
Additionally, we also report the mF1* and mPQ* scores which are solely computed on new classes in the target domain.

\noindent\textbf{Cancer Tissue Phenotyping} \ 
In this setting, we perform adaptation from a two-class discrimination task to an eight-class one on CRC-TP \cite{javed2020cellular} and Kather \cite{kather2016multi} datasets.
Kather consists of 5000 $150\times 150$ pathology image patches and includes six novel classes other than the two common classes also existing in the source domain.
The experiments are conducted with ResNet-101 as backbone.
For evaluation, we exploit the accuracy and F1 scores as metrics to indicate the classification performance.
The average scores for all classes\;(mAcc and mF1) and particularly the ones on new classes only\;(mAcc* and mF1*) are reported.

\noindent\textbf{Skin Lesion Diagnosis} \ 
We construct a fine-grained taxonomy adaptation setting from two coarse classes to seven subclasses with HAM10000 \cite{tschandl2018ham10000}, which contains 10015 dermatoscopic images sampled from different body structures.
The adopted experiment settings and evaluation metrics are the same as the cancer tissue phenotyping task.
Since the label space of the source and target datasets do not overlap, we only report the mAcc* and mF1* scores which are computed on target classes.

Following previous works in UDA \cite{chen2019synergistic}, for all experiments, each dataset is randomly split into $80\%/20\%$ for training/testing.
To explicitly recognize the target-private new classes, we sample few\;(5/10) samples with corresponding labels to formulate the support set.
The remaining target data is left unlabeled.
More details can be found in the supplemental material.

\begin{table*}[!t]
	\centering
	\fontsize{8.5}{9.5}\selectfont
	\begin{threeparttable}
		\caption{Comparison results of our proposed method against other state-of-the-art methods for cancer tissue phenotyping. 
		}
		\label{tab:tissue_result}
		\setlength{\tabcolsep}{2.3mm}{
			\begin{tabular}{ccccccccc}
				\toprule[0.5mm]
				\multirow{3}{*}{\footnotesize \textbf{Methods}}
				&\multicolumn{4}{c}{\textbf{5-shot}}
				&\multicolumn{4}{c}{\textbf{10-shot}}\cr
				\cmidrule(lr){2-5} \cmidrule(lr){6-9} 
				&mAcc&mAcc*&mF1&mF1*&mAcc&mAcc*&mF1&mF1*\cr
				\midrule[0.3mm]				
				Sup-only
				&46.82$_{1.30}$&42.26$_{1.26}$&44.64$_{1.49}$&42.83$_{1.08}$
				&48.52$_{1.02}$&45.93$_{1.67}$&45.34$_{1.45}$&46.15$_{1.83}$\cr
				Multi-task\,\cite{sener2018multi}
				&48.35$_{0.78}$&42.02$_{1.34}$&45.55$_{1.12}$&41.68$_{1.63}$
				&48.56$_{0.69}$&45.00$_{1.72}$&46.03$_{0.64}$&44.46$_{1.21}$\cr
				\midrule[0.2mm]			
				DANN\,\cite{ganin2016domain}
				&47.46$_{0.67}$&42.53$_{1.22}$&43.08$_{1.14}$&43.13$_{1.30}$
				&48.42$_{0.48}$&45.31$_{1.27}$&46.50$_{0.81}$&44.02$_{0.64}$\cr
				CGDM\,\cite{du2021cross}
				&49.02$_{0.82}$&\textcolor{brown}{43.53$_{1.56}$}&\textcolor{brown}{46.29$_{0.87}$}&\textcolor{brown}{43.58$_{0.69}$}
				&\textcolor{brown}{50.54$_{0.84}$}&44.57$_{1.35}$&\textcolor{brown}{48.17$_{0.66}$}&45.23$_{0.40}$\cr
				\midrule[0.2mm]						
				LETR\,\cite{luo2017label}
				&45.47$_{1.95}$&41.33$_{1.47}$&43.17$_{1.88}$&41.20$_{2.02}$
				&45.94$_{0.56}$&40.13$_{1.30}$&41.16$_{1.13}$&39.61$_{0.90}$\cr
				FT-CIDA\,\cite{kundu2020class}
				&43.72$_{0.88}$&39.06$_{0.79}$&40.91$_{1.02}$&38.73$_{0.84}$
				&44.65$_{0.53}$&38.80$_{0.94}$&40.94$_{0.96}$&38.96$_{1.07}$\cr
				STARTUP\,\cite{phoo2021selftraining}
				&\textcolor{brown}{49.52$_{0.63}$}&42.93$_{1.83}$&43.59$_{1.37}$&39.97$_{0.59}$
				&49.40$_{0.51}$&\textcolor{brown}{46.91$_{1.87}$}&46.38$_{0.70}$&\textcolor{brown}{47.85$_{1.36}$}\cr	
				DDN\,\cite{islam2021dynamic}
				&47.14$_{1.75}$&43.25$_{0.99}$&42.92$_{2.29}$&40.76$_{0.50}$
				&48.27$_{0.67}$&44.08$_{1.22}$&46.03$_{1.14}$&45.44$_{1.18}$\cr	
				TSA\,\cite{li2022crossfew}
				&48.22$_{1.49}$&42.13$_{0.78}$&44.39$_{1.84}$&41.92$_{1.13}$
				&47.18$_{1.02}$&41.26$_{1.10}$&45.34$_{1.72}$&44.69$_{1.89}$\cr
				TACS\,\cite{gong2022tacs}
				&45.30$_{1.07}$&40.28$_{1.11}$&44.74$_{1.26}$&42.04$_{0.95}$
				&46.95$_{1.16}$&43.69$_{1.74}$&47.46$_{1.51}$&47.51$_{1.25}$\cr	
				\rowcolor{light-gray} Ours
				&\textbf{50.14}$_{0.76}$&\textbf{47.66}$_{1.54}$&\textbf{47.41}$_{1.10}$&\textbf{46.45}$_{1.15}$
				&\textbf{55.86}$_{0.98}$&\textbf{51.27}$_{1.02}$&\textbf{52.98}$_{0.49}$&\textbf{52.12}$_{0.94}$\cr	
				\bottomrule[0.5mm]
		\end{tabular}}
	\end{threeparttable}
\end{table*}

\begin{table}[!t]
	\centering
	\fontsize{8.5}{9.5}\selectfont
	\begin{threeparttable}
		\caption{Comparison results of our proposed method against other state-of-the-art methods for skin lesion diagnosis. 
		}
		\label{tab:skin_result}
		\setlength{\tabcolsep}{1.2mm}{
			\begin{tabular}{ccccc}
				\toprule[0.5mm]
				\multirow{3}{*}{\footnotesize \textbf{Methods}}
				&\multicolumn{2}{c}{\textbf{5-shot}}
				&\multicolumn{2}{c}{\textbf{10-shot}}\cr
				\cmidrule(lr){2-3} \cmidrule(lr){4-5} 
				&mAcc*&mF1*&mAcc*&mF1*\cr
				\midrule[0.3mm]				
				Sup-only
				&35.71$_{0.79}$&19.89$_{0.22}$
				&38.50$_{0.38}$&22.36$_{0.67}$\cr
				Multi-task\,\cite{sener2018multi}
				&33.80$_{1.20}$&19.06$_{0.31}$
				&37.85$_{0.40}$&23.32$_{0.32}$\cr
				\midrule[0.2mm]			
				DANN\,\cite{ganin2016domain}
				&37.22$_{0.57}$&23.99$_{0.92}$
				&40.06$_{0.94}$&24.32$_{0.72}$\cr
				CGDM\,\cite{du2021cross}
				&41.50$_{1.03}$&24.26$_{0.80}$
				&\textcolor{brown}{47.22$_{0.70}$}&24.00$_{0.23}$\cr
				\midrule[0.2mm]						
				LETR\,\cite{luo2017label}
				&36.04$_{0.42}$&21.39$_{0.66}$
				&40.87$_{1.22}$&23.50$_{0.49}$\cr
				FT-CIDA\,\cite{kundu2020class}
				&31.95$_{0.77}$&18.13$_{0.48}$
				&34.60$_{0.90}$&20.96$_{0.88}$\cr
				STARTUP\,\cite{phoo2021selftraining}
				&43.21$_{1.15}$&\textcolor{brown}{27.17$_{0.92}$}
				&43.88$_{1.35}$&25.59$_{0.60}$\cr
				DDN\,\cite{islam2021dynamic}
				&36.19$_{1.05}$&21.50$_{0.34}$
				&42.46$_{1.29}$&23.92$_{0.28}$\cr
				TSA\,\cite{li2022crossfew}
				&40.41$_{1.36}$&24.05$_{1.01}$
				&42.07$_{1.51}$&25.44$_{1.06}$\cr
				TACS\,\cite{gong2022tacs}
				&\textcolor{brown}{45.05$_{0.46}$}&25.60$_{0.65}$
				&46.73$_{0.64}$&\textcolor{brown}{25.96$_{0.79}$}\cr	
				\rowcolor{light-gray} Ours
				&\textbf{50.79}$_{0.69}$&\textbf{30.05}$_{0.81}$&\textbf{52.14}$_{0.98}$&\textbf{30.52}$_{0.83}$\cr	
				\bottomrule[0.5mm]
		\end{tabular}}
	\end{threeparttable}
\end{table}

\subsection{Results}
We compare our method with state-of-the-art approaches for cross-domain adaptation, including UDA methods DANN \cite{ganin2016domain} and CGDM \cite{du2021cross} which do not consider the taxonomic inconsistency, as well as LETR \cite{luo2017label}, FT-CIDA\cite{kundu2020class}, STARTUP \cite{phoo2021selftraining}, DDN \cite{islam2021dynamic}, TSA \cite{li2022crossfew}, and TACS \cite{gong2022tacs} which are designed for both domain shifts and new class learning.
For UDA methods, we jointly perform domain alignment and train the model with the labeled target data.
In addition, we conduct comparisons with multi-task learning approach \cite{sener2018multi}, which trains the model on all domains and label sets simultaneously.
``Sup-only'' indicates the baseline where only the supervised loss on the labeled target dataset is used for training.

Table\;\ref{tab:nuclei_result} shows the comparison results on the cross-domain nuclei recognition and segmentation task.
All methods adopt the same ResNet-50-HoverNet as base model following \cite{graham2019hover}.
It can be observed that our proposed method outperforms the state-of-the-art cross-domain adaptation approaches by a large margin.
We also notice an important point that compared with the ``Sup-only'' baseline, most competing methods fail to attain better performance on the metrics specific to new classes\;(\emph{i.e.,} mF1* and mPQ*).
For example, under the 5-shot setting, despite a 4.68$\%$ improvement is achieved by DANN on mF1, it achieves inferior performance on mF1* which is 4.94$\%$ worse than ``Sup-only''.
This phenomenon is incurred by the failure of cross-class generalization.
In contrast, our method yields substantial improvements on those new class-specific metrics,
which indicates the superior generalizability of our method.
The observation is further verified through visual comparisons in the supplementary material.

The overall quantitative results on the cancer tissue phenotyping and skin lesion diagnosis tasks are presented in Table\;\ref{tab:tissue_result} and \ref{tab:skin_result}, respectively.
All methods are implemented based on the ResNet-101 backbone.
Under each few-shot setting, our method shows higher classification scores compared with previous approaches.
In particular, we achieve 0.62$\%$$\sim$5.32$\%$ and 1.12$\%$$\sim$4.81$\%$ improvements against the second-best results in terms of mAcc and mF1 for cancer tissue phenotyping,
as well as 4.92$\%$$\sim$7.58$\%$ and 2.88$\%$$\sim$4.93$\%$ improvements in terms of mAcc* and mF1* for skin lesion diagnosis.
The higher accuracy consistently attained by our method under different cross-domain adaptation settings and clinical tasks is attributed to the strong generalizability brought by the proposed optimization trajectory distillation approach.

\subsection{Analysis and Discussion}
\noindent\textbf{Ablation of Key Components} \ 
Table\;\ref{tab:ablation_joint} shows the ablation study results to illustrate the effectiveness of key components.
``projection'' denotes the gradient projection approach introduced in Section\;\ref{sec:cross-distill}.
For the ``\emph{w/o} projection'' ablation, we perform cross-domain and cross-class distillation by directly minimizing the statistics discrepancy between unrectified gradients $(g_s,g_t)$ and $(g_A,g_N)$ in Eq.\;(\ref{equ:project}).
It is observed that each component is indispensable and could improve the overall performance, which elaborates the importance of external navigation for learning target domain and new classes in network training.
Interestingly, we notice that the impacts of cross-domain and cross-class distillation could be disparate across tasks.
For example, cross-class distillation brings higher performance gain compared with cross-domain distillation on the cancer tissue phenotyping benchmark, while the situation is reversed for skin lesion diagnosis.
It is caused by the different natures of each task.
Class-imbalance is a serious issue for skin lesion diagnosis in that the class distribution is highly biased \cite{tschandl2018ham10000}, which compromises the effectiveness of the cross-class distillation module.
Differently, for cancer tissue phenotyping, classes are uniformly distributed \cite{kather2016multi} and could hence ensure the representativity of anchor classes and their exemplarity to be distilled to new classes.

\begin{table*}[!t]
	\centering
	\fontsize{8.5}{9.5}\selectfont
	\begin{threeparttable}
		\caption{Key component analysis of our method on three cross-domain adaptation benchmarks under 10-shot scheme. The best results are highlighted in bold.
		}
		\label{tab:ablation_joint}
		\setlength{\tabcolsep}{1.7mm}{
			\begin{tabular}{ccccccccccc}
				\toprule[0.5mm]
				\multirow{3}{*}{\textbf{Methods}}
				&\multicolumn{4}{c}{\textbf{Nuclei}}
				&\multicolumn{4}{c}{\textbf{Cancer}}
				&\multicolumn{2}{c}{\textbf{Skin}}\cr
				\cmidrule(lr){2-5} \cmidrule(lr){6-9} \cmidrule(lr){10-11}
				&mF1&mF1*&mPQ&mPQ*&mAcc&mAcc*&mF1&mF1*&mAcc*&mF1*\cr
				\midrule[0.3mm]				
				Full Framework
				&\textbf{43.88}&\textbf{31.43}&\textbf{24.81}&\textbf{19.35}
				&\textbf{55.86}&\textbf{51.27}&\textbf{52.98}&\textbf{52.12}
				&\textbf{52.14}&\textbf{30.52}\cr	
				\emph{w/o} cross-domain
				&39.96&30.08&22.29&18.10
				&52.81&49.93&51.62&50.64
				&44.34&25.28\cr	
				\emph{w/o} cross-class
				&42.30&27.82&23.76&17.05
				&48.90&47.22&49.85&44.76
				&49.06&28.24\cr		
				\emph{w/o} historical
				&41.54&30.47&23.00&17.59
				&50.17&48.05&48.19&45.12
				&45.83&25.97\cr	
				\emph{w/o} projection
				&40.06&28.80&22.10&17.31
				&53.58&45.78&49.01&49.20
				&47.29&28.02\cr	
				\bottomrule[0.5mm]
		\end{tabular}}
	\end{threeparttable}
\end{table*}

\begin{figure}[!t]
	\centerline{\includegraphics[width=0.98\columnwidth]{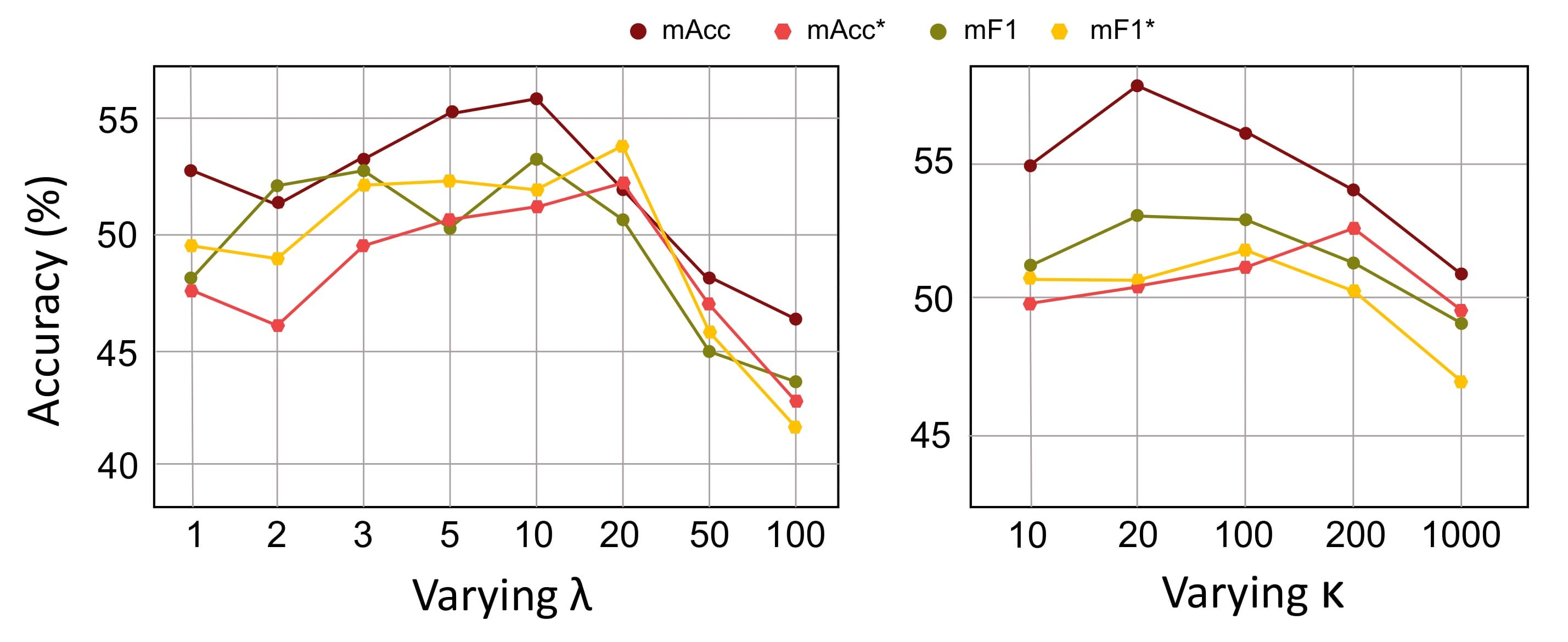}}
	\caption{Performance comparison between different choices of hyperparameters $\lambda$ and $\kappa$ on the cancer tissue phenotyping benchmark under 10-shot scheme.}
	\label{fig:parameter_analysis}
\end{figure}

\noindent\textbf{Hyperparameter Sensitivity} \ 
To investigate the effect of hyperparameters $\lambda$ in Eq.\;(\ref{equ:objective}) and $\kappa$ in Eq.\;(\ref{equ:step}), we present the performance comparison by varying the choices of their values for cancer tissue phenotyping, as shown in Fig.\;\ref{fig:parameter_analysis}.
We only change one hyperparameter while keeping the other fixed at a time.
The experimental results show that our framework is quite stable for hyperparameters in a wide interval, yet setting $\lambda$ and $\kappa$ to a large value is detrimental to the classification accuracy.
Specifically, $\lambda$ is a balancing coefficient between class discriminativeness and gradient alignment.
When it is too large, the trained model is biased to matching learning dynamics and would lose essential semantic knowledge.
For $\kappa$, it controls the proportions of unrectified gradients.
Its large value indicates that most original gradients are preserved, which would incur noisy optimization steps and loss of flatness.
Therefore, we decide to set $\lambda=10$ and $\kappa=100$.
It achieves the best performance under most metrics.

\begin{figure}[!t]
	\centerline{\includegraphics[width=\columnwidth]{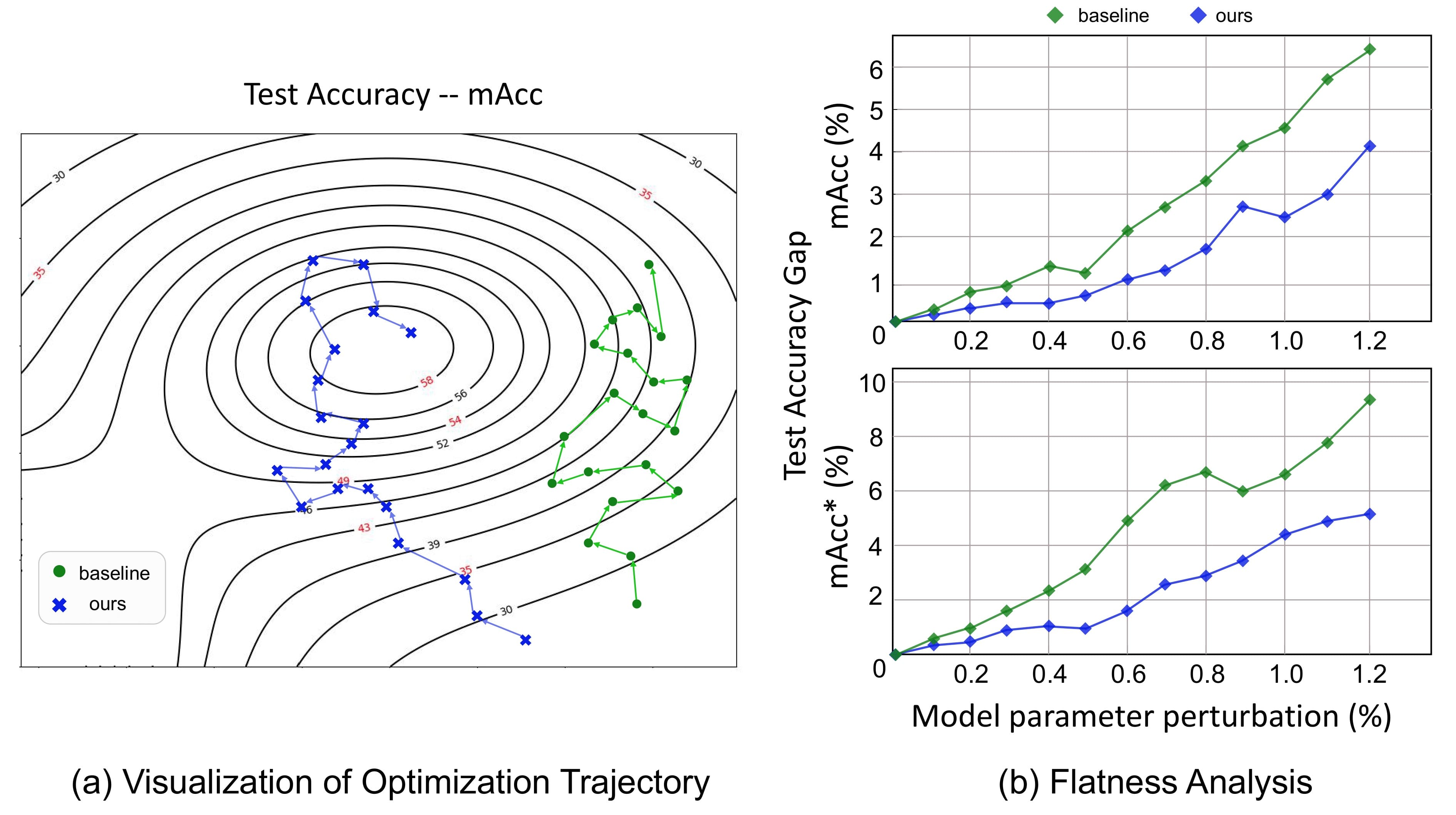}}
	\caption{Further analysis of our proposed method. 
		(a)\;We plot the optimization trajectory of our method and the multi-task baseline on the test accuracy surface. 
		(b)\;We visualize the local flatness of our method and the baseline via test accuracy gap by varying the model parameter perturbation rate. Each point is computed by Monte-Carlo approximation over 50 random samples.
		All the experiments are conducted on the cancer tissue phenotyping benchmark under 10-shot scheme.}
	\label{fig:vis_OT_flat}
\end{figure}

\noindent\textbf{Impacts on Optimization Trajectory and Model Flatness} \ 
To better understand the effectiveness of our method, we perform in-depth analysis about its impacts on model's training trajectory and flatness of the converged local minima.
Firstly, as shown in Fig.\;\ref{fig:vis_OT_flat}(a), we visualize the optimization paths of our method and the multi-task baseline on the test accuracy surface with a contour plot. The details of plot procedure can refer to \cite{izmailov2018averaging}.
It can be observed that the optimization of the multi-task baseline fluctuates sharply and is restricted in a sub-optimal area with a relatively low test accuracy.
This is induced by the biased navigation in the course of training due to domain shifts and category gap.
With the devised dual-stream optimization trajectory distillation module, our method could by contrast suppress the noisy update signals and proceed consistently towards the optimum.

Besides, we perform analysis about the solutions found by our method and the baseline in terms of flatness \cite{cha2021swad}.
Specifically, the local flatness of a model parameter $\theta$ is quantified by the expected changes of test accuracy between $\theta$ and a neighbouring $\theta^*$ with perturbation rate $\varrho$: $\mathcal{F}_\varrho(\theta)=\mathbb{E}_{\left\|\theta^*\right\|=(1+\varrho)\left\|\theta\right\|}\big[\rm{Acc}(\theta^*) - \rm{Acc}(\theta)\big]$.
We approximate the expected value by Monte-Carlo sampling over 50 samples.
As demonstrated in Fig.\;\ref{fig:vis_OT_flat}(b), comparing with the baseline approach, our method shows stronger robustness against model parameter perturbation.
It proves that our method successfully converges to minima with better local flatness and thus possesses superior generalizability \cite{keskar2017on}.

\section{Conclusion}
In this paper, we study the taxonomy adaptive cross-domain adaptation paradigm, which allows incoherent label sets between source and target datasets.
To jointly address the data distribution bias and category gap, we propose a unified framework which performs cross-domain and cross-class optimization trajectory distillation to calibrate the learning dynamics of insufficiently annotated domain and classes.
Historical self-distillation is further devised to attain a well-generalizable solution via 
regulating gradient distributions.
Extensive experiments on multiple benchmarks with different task contexts demonstrate the effectiveness and generalization capability of our method.
In the future, we will extend this work to further discover the underlying mechanism of optimization trajectory distillation for model generalization and evaluate its potential in more adaptation scenarios.


\clearpage
\title{Supplementary Material: Taxonomy Adaptive Cross-Domain Adaptation in Medical Imaging via Optimization Trajectory Distillation}

\maketitle
\setcounter{section}{0}

In this supplementary, we provide additional illustration of our proposed method and proofs to support theoretical analysis, 
as well as dataset and implementation details.
Extended experiments and analysis are performed to further verify the effectiveness and robustness of our method.

\section{Overall Traininig Procedure}
\begin{algorithm}[!h]
	\label{alg:1}
	\begin{small}
		\caption{Training Procedure for Optimization Trajectory Distillation}\label{algorithm}
		\KwIn{Source dataset $\mathcal{D}_s$, target dataset $\mathcal{D}_t$, number of iterations $I$}
		\KwOut{Optimized model parameters $\theta$}
		$\textbf{Init:}$ Gradient memory buffer $\bm{G}_s,\bm{G}_A,\bm{G}_{It}$\\
		\For{$i\leftarrow 1$ \KwTo $I$}{
			$\{(\mathbf{x}_s, \mathbf{y}_s)\}\leftarrow$ Image-label pairs sampled from$(\mathcal{D}_s)$\\
			$\{(\mathbf{x}_t^u)\}, \{(\mathbf{x}_t^l, \mathbf{y}_t^l)\}\leftarrow$ Sample from$(\mathcal{D}_t)$\\
			Generate online pseudo-label $\tilde{\mathbf{y}}_t^u$\\
			$\{(\mathbf{x}_t, \mathbf{y}_t)\}\leftarrow$ $\{(\mathbf{x}_t^u, \tilde{\mathbf{y}}_t^u), (\mathbf{x}_t^l, \mathbf{y}_t^l)\}$\\
			Backpropagate gradients $\bm{g}_s,\bm{g}_t,\bm{g}_A,\bm{g}_N$ by Eq.(2)\\
			$\textcolor{brown}{\textbf{Cross-domain/class distillation:}}$\\
			Update $\bm{G}_s$ and $\bm{G}_A$ with $\bm{g}_s$ and $\bm{g}_A$\\
			\If{$\bm{G}_s$ $\rm{and}$ $\bm{G}_A$ $\rm{are\ full}$}
			{Apply SVD to identify the principal subspace and form the corresponding projection matrix $\bm{M}_s,\bm{M}_A$ by Eq.(5)(6)\\
				Clear $\bm{G}_s$ and $\bm{G}_A$
			}
			Perform gradient projection by Eq.(7)\\
			Compute the overall training objective $\mathcal{L}$ by Eq.(8)\\
			$\textcolor{brown}{\textbf{Temporal self-distillation:}}$\\
			Update $\bm{G}_{It}$ with $\bm{g}_{It}$\\
			\If{$\bm{G}_{It}$ $\rm{is\ full}$}
			{Form the projection matrix $\bm{M}_{It}$ by Eq.(5)(6)\\
				Clear $\bm{G}_{It}$
			}
			Compute the mini-batch gradients $\tilde{\bm{g}}\leftarrow\nabla_\theta\mathcal{L}(\theta)$\\
			Update model parameters $\theta$ by Eq.(11)
		}
		$\textbf{return}$ $\theta$ 
	\end{small}
\end{algorithm}
We summarize the workflow of our proposed method in Algorithm\;\ref{alg:1}.
The referred equation can be found in the main text.

\section{Details for Theoretical Analysis}
\subsection{Joint Characterization of Feature and Output Space}
Here, we prove that in our gradient-based method, information in feature space and model-output space can be jointly modeled implicitly, which demonstrates the superiority of our method as a unified framework that jointly characterizes the feature and output space as well as the learning dynamics.
We illustrate this property under cross-entropy loss and Dice loss.

Specifically, for the task-specific prediction layer, which is implemented as a convolutional layer with $1\times1$ kernel, its function can be mathematically expressed as:
\begin{equation}
	\bm{u} = \bm{W}^\top\bm{z} + \bm{b},
\end{equation}%
where $\bm{z}\in \mathbb{R}^{B\times m\times h\times w}$ denotes the input feature maps with batch size $B$, channel number $m$, and spatial size $h\times w$. 
$\bm{W}\in \mathbb{R}^{m \times c}$ is the convolutional kernel matrix with input channel $m$ and output channel $c$.
$\bm{b}\in \mathbb{R}^{c}$ indicates the bias tensor.
$\bm{u}\in \mathbb{R}^{B\times c\times h\times w}$ is the logit predictions.

We first consider the gradients for network parameter $\theta$ by backpropagating the cross entropy\;(CE) loss $\mathcal{L}_{CE}$:
\begin{equation}
	\min_{\theta \sim [\bm{W},\,\bm{b}]}\,-\sum_{n=1}^{p}\sum_{i=1}^{c} \bm{y}_i^n\log\frac{e^{\bm{u}_i^n}}{\sum_{j=1}^c e^{\bm{u}_j^n}},
\end{equation}%
where $p = B\times h\times w$ is the total number of pixels, $\bm{y}\in \mathbb{R}^{B\times c\times h\times w}$ is the one-hot pixel-wise class label.
Since our focus is on feature space $\bm{z}$ and output space $\bm{u}$, without loss of generality, we set $\bm{y}$ to follow the uniform class distribution that $\bm{y} = [1/c,\,1/c, ..., 1/c]$.
Then, for each pixel, the derivative of the CE loss can be formulated as:
\begin{equation}
	\begin{split}
		&\frac{\partial \mathcal{L}_{CE}}{\partial \theta}\bigg|_{\theta = [\theta_1,\,\theta_2,\,...,\,\theta_c]}
		=\frac{\partial \mathcal{L}_{CE}}{\partial \bm{u}}\bigg|_{\bm{u} = [\bm{u}_1,\,\bm{u}_2,\,...,\,\bm{u}_c]}\cdot\frac{\partial \bm{u}}{\partial \theta} \\
		&=-\frac{1}{c}\cdot\frac{\partial\,(\sum_{i=1}^{c}\log\frac{e^{\bm{u}_{i}}}{\sum_{j=1}^c e^{\bm{u}_{j}}})}{\partial \bm{u}}\cdot\frac{\partial \bm{u}}{\partial \theta} \\
		&=-\frac{1}{c}\cdot\frac{\partial\,(\sum_{i=1}^{c}\bm{u}_{i}-c\cdot\log{\sum_{j=1}^c e^{\bm{u}_{j}}})}{\partial \bm{u}}\cdot\frac{\partial \bm{u}}{\partial \theta} \\
		&=-\frac{1}{c}\cdot(1 - c\cdot\frac{[e^{\bm{u}_{1}},\,e^{\bm{u}_{2}},\,...,\,e^{\bm{u}_{c}}]}{\sum_{j=1}^c e^{\bm{u}_{j}}})\cdot\bm{z}.
	\end{split}
\end{equation}%
Then we summarize the gradient magnitudes over all pixels and channels:
\begin{equation}
	\begin{split}
		&\sum_{i=1}^{c}\lVert\frac{\partial \mathcal{L}_{CE}}{\partial \theta_i}\rVert =
		\sum_{n=1}^{p}\sum_{i=1}^{c}\lVert\frac{\partial \mathcal{L}_{CE}}{\partial \bm{u}_i^n}\cdot\frac{\partial \bm{u}_i^n}{\partial \theta_i}\rVert \\
		&=-\frac{1}{c}\cdot(\underbrace{\sum_{n=1}^{p}\sum_{i=1}^{c}\lVert1 - c\cdot\frac{e^{\bm{u}_i^n}}{\sum_{j=1}^c e^{\bm{u}_j^n}}}_{\mathrm{output\,space}}\rVert)\cdot\underbrace{(\sum_{n=1}^{p}\lVert\bm{z}^n\rVert)}_{\mathrm{feature\,space}}.
	\end{split}
\end{equation}%
The result indicates that the gradients of CE loss characterize the information in both feature space and output space.

Similarly, for Dice loss $\mathcal{L}_{Dice}$:
\begin{equation}
	\min_{\theta \sim [\bm{W},\,\bm{b}]}\,\sum_{n=1}^{p}\sum_{i=1}^{c}(1-
	\frac{2\bm{y}_i^n\mathtt{act}(\bm{u}_i^n)}
	{\bm{y}_i^n+\mathtt{act}(\bm{u}_i^n)}).
\end{equation}%
Here we omit the details of softmax layer and use $\mathtt{act}$ to denote the activation function for simplicity.
When $c$ is large, the pixel-wise derivative can be approximated by:
\begin{equation}
	\begin{split}
		&\frac{\partial \mathcal{L}_{Dice}}{\partial \theta}\bigg|_{\theta = [\theta_1,\,\theta_2,\,...,\,\theta_c]} \\
		&=\frac{\partial \mathcal{L}_{Dice}}{\partial \mathtt{act}(\bm{u})}\cdot\frac{\partial \mathtt{act}(\bm{u})}{\partial \bm{u}}\cdot\frac{\partial \bm{u}}{\partial \theta} \\
		&\approx\bm{\xi}\cdot\frac{[e^{\bm{u}_{1}},\,e^{\bm{u}_{2}},\,...,\,e^{\bm{u}_{c}}]}{[\mathtt{act}(\bm{u}_{1})^2,\,\mathtt{act}(\bm{u}_{2})^2,\,...,\,\mathtt{act}(\bm{u}_{c})^2]}\cdot\bm{z},
	\end{split}
\end{equation}%
where $\bm{\xi}$ is a constant term.
Its gradient magnitudes can be written as:
\begin{equation}
	\begin{split}
		\sum_{i=1}^{c}\lVert\frac{\partial \mathcal{L}_{Dice}}{\partial \theta_i}\rVert &=
		\sum_{n=1}^{p}\sum_{i=1}^{c}\lVert\frac{\partial \mathcal{L}_{Dice}}{\partial \mathtt{act}(\bm{u}_i^n)}\cdot\frac{\partial \mathtt{act}(\bm{u}_i^n)}{\partial \bm{u}_i^n}\cdot\frac{\partial \bm{u}_i^n}{\partial \theta_i}\rVert \\
		&\approx\bm{\xi}\cdot(\underbrace{\sum_{n=1}^{p}\sum_{i=1}^{c}\lVert\frac{e^{\bm{u}_i^n}}{\mathtt{act}(\bm{u}_i^n)^2}}_{\mathrm{output\,space}}\rVert)\cdot\underbrace{(\sum_{n=1}^{p}\lVert\bm{z}^n\rVert)}_{\mathrm{feature\,space}},
	\end{split}
\end{equation}%
which proves that the proposition also holds true for Dice loss.

\subsection{Impacts on Generalization Error}
In this section, we prove the effectiveness of our method towards a tighter generalization error bound on the target domain and novel classes.

Firstly, we analyze the underlying mechanism for the cross-domain distillation module.
Let $\mathcal{H}$ be a hypothesis space of VC-dimension $d$, for $h\in \mathcal{H}$, the correlations between the error on the target domain and the distance in gradient space across domains are established as \cite{ben2006analysis, gao2021gradient}:
\begin{equation}\begin{split}
		\epsilon&_T(h)\leq\hat{\epsilon}_S(h)+\frac{4}{n_l}\sqrt{(d\log\frac{2en_l}{d}+\log\frac{4}{\delta})}+\Lambda \\
		&+\mathbf{Div}_\nabla\boldsymbol{(\tilde{\mathcal{U}}_S,\tilde{\mathcal{U}}_T)}+4\sqrt{\frac{4\log(2n_u)+\log(\frac{4}{\delta})}{n_u}},
\end{split}\end{equation}%
with probability at least $1-\delta$.
Here $\epsilon_T$ and $\hat{\epsilon}_S$ represent the true and empirically estimated error of the target and source domain, respectively. 
$Div_\nabla(\tilde{\mathcal{U}}_S,\tilde{\mathcal{U}}_T)$ is the distance between data distributions $\tilde{\mathcal{U}}_S$ and $\tilde{\mathcal{U}}_T$ in gradient space.
$n_l$ and $n_u$ denote the number of labeled and unlabeled samples.
$\Lambda$, $\delta$, and $e$ are constants.
It implies that constraining the gradient descent trajectory of the target domain to approximate the source domain's, which reduces $Div_\nabla(\tilde{\mathcal{U}}_S,\tilde{\mathcal{U}}_T)$, could lead to lower cross-domain generalization error.

Furthermore, we demonstrate that the cross-class distillation module contributes to lower empirical error on novel classes from the multi-task learning perspective.
Suppose that $\mathcal{L}$ is the empirical training loss and $\nabla_\theta\mathcal{L}_q(\theta)$ denotes its derivative \emph{w.r.t.} class $q$.
Given a set of anchor classes $\{a_i\}_{i=1}^A$ and a novel class $q$, with the first-order Taylor expansion, we have:
\begin{equation}
	\mathcal{L}_q(\theta-\mu\cdot\Delta\theta^*) = \mathcal{L}_q(\theta) - \mu\cdot\nabla_\theta\mathcal{L}_q(\theta)\cdot\Delta\theta^* + \mathcal{O}(\mu),
\end{equation}%
where the optimization step $\Delta\theta^*$ is characterized by $\sum_{i=1}^{A}\nabla_\theta\mathcal{L}_{a_i}(\theta) + \nabla_\theta\mathcal{L}_q(\theta)$, $\mu$ is a small value. 
Then:
\begin{equation}\begin{split}
		\mathcal{L}_q(\theta-\mu\cdot&\Delta\theta^*) - \mathcal{L}_q(\theta) = -\mu\cdot\Big{\{}\left\|\nabla_\theta\mathcal{L}_q(\theta)\right\|^2 \cr
		+&\sum_{i=1}^{A}\big{[}\boldsymbol{\nabla_\theta\mathcal{L}_q(\theta)\cdot\nabla_\theta\mathcal{L}_{a_i}(\theta)}\big{]}\Big{\}} + \mathcal{O}(\mu).
\end{split}\end{equation}%
It indicates that by enforcing the similarity between the gradients \emph{w.r.t.} novel and anchor classes, we could drive the model to reduce the empirical loss on novel classes along optimization and thereby attain a well-generalizable solution.

\section{Implementation Details}
\subsection{Nuclei Segmentation and Recognition}
\subsubsection{Datasets and Preprocessing}
Accurate detection, segmentation, and classification of nuclei serve as essential prerequisites for various clinical and research studies within the digital pathology field \cite{graham2019hover}.
Inconsistent taxonomy for nuclei categorization is common across different institutes, which results in the unmatched label sets among datasets.
In this regard, we use PanNuke \cite{gamper2019pannuke} and Lizard \cite{graham2021lizard} as the source and target dataset, respectively.
\textbf{PanNuke} contains 481 visual fields cropped from whole-slide images along with 189,744 annotated nuclei. 
It follows a categorization schema where nuclei are divided into five classes, including neoplastic, non-neoplastic epithelial, inflammatory, connective, and dead cells.
We discard the ``dead'' class as it does not exist in most image patches.
To ensure the dataset has a uniform data distribution, we use all images from the breast tissue to formulate the source dataset.
\textbf{Lizard} consists of 291 image regions with an average size of $1016\times 917$ pixels from the colon tissue and annotates 495,179 nuclei.
It adopts a categorization schema different to PanNuke that there are six classes in total, \emph{i.e.}, neutrophil, eosinophil, plasma, lymphocyte, epithelial, and connective cells.
We use the Dpath subset as the target dataset.
For preprocessing, all the visual fields with divergent size are randomly cropped into image patches of $128\times 128$ pixels.
CutMix \cite{yun2019cutmix} is used to augment the target dataset.

\subsubsection{Network Architectures and Parameter Settings}
We employ the widely used Hover-Net \cite{graham2019hover} architecture with a standard ResNet-50 backbone as the base model.
The optimizer is Adam with a learning rate of $1e-4$ and $(\beta_1,\beta_2)=(0.9,0.999)$, and the batch size is set as 4.
To supervise the classification and segmentation branches, we adopt a combined loss of $\rm{CrossEntropyLoss+ DiceLoss}$.
$\lambda$ in Eq.(8) and $\kappa$ in Eq.(11) are empirically set to 1000 and 10, respectively.

\subsubsection{Evaluation Metrics}
F1 score is a popular metric to evaluate classification performance.
It measures both precision and recall harmonically.
We report the class-averaged score to indicate the overall accuracy.
Panoptic quality\;(PQ) \cite{graham2019hover} is a unified metric for the instance segmentation task which models the quality of both detection and segmentation results concurrently:
\begin{equation}
	\mathcal{PQ} = \underbrace{\frac{\rm{TP}}{\rm{TP}+\frac{1}{2}\rm{FP}+\frac{1}{2}\rm{FN}}}_{\rm{Detection\  Quality}}\cdot\underbrace{\frac{\sum_{(y, \hat{y})\in\rm{TP}}IoU(y, \hat{y})}{\rm{TP}}}_{\rm{Segmentation\  Quality}}.
\end{equation}
where $\rm{TP, FP, FN}$ are the true positive, false positive, and false negative detection predictions, respectively.
$(y, \hat{y})$ represents the pair of ground truth and predicted segmentation mask.
$IoU$ is the intersection over union score.

\subsection{Cancer Tissue Phenotyping}
\subsubsection{Datasets and Preprocessing}
Identifying distinct tissue phenotypes is an essential step towards systematic profiling of the tumor microenvironment in pathological examination \cite{javed2020cellular}.
Previous works are mostly limited to the discrimination of two classes of tissue: tumor and stroma \cite{linder2012identification}, while recent studies argue that recognizing more heterogeneous tissues brings clinical value \cite{kather2016multi}.
We therefore propose to perform adaptation from a dataset with only two categories of tissue to another dataset with several novel classes.
In particular, we select images of tumor and stroma tissue from the CRC-TP \cite{javed2020cellular} to form the source dataset.
\textbf{CRC-TP} contains 20 H$\rm{\&}$E-stained colorectal cancer\;(CRC) slides obtained from 20 different patients.
Region-level tissue phenotype annotations are provided by expert pathologists.
To ensure a unique category label can be assigned to each image,
patches are extracted at $20\times$ magnification with the size of $150\times 150$ pixels.
The patch-wise tissue phenotypes are decided based on the majority of their content.
\textbf{Kather} \cite{kather2016multi} is then regarded as the target dataset.
It consists of 5000 $150\times 150$ pathology image patches sampled from 10 anonymized CRC tissue slides.
Other than tumor and stroma tissue, Kather includes six novel tissue types, \emph{i.e.}, complex stroma, immune cells, debris, normal mucosal glands, adipose tissue, background, and thus poses an 8-class classification problem.

\subsubsection{Network Architectures and Parameter Settings}
For experiments, we employ ResNet-101 as the image encoder and thereupon add two classification heads on top to perform 2-class and 8-class discrimination, respectively.
During training, cross-entropy loss and Adam optimizer with learning rate $1e-4$ are used to optimize the model with a batch size of 4.
$\lambda$ and $\kappa$ are set to 10 and 100.

\begin{figure*}[!t]
	\centerline{\includegraphics[width=2\columnwidth]{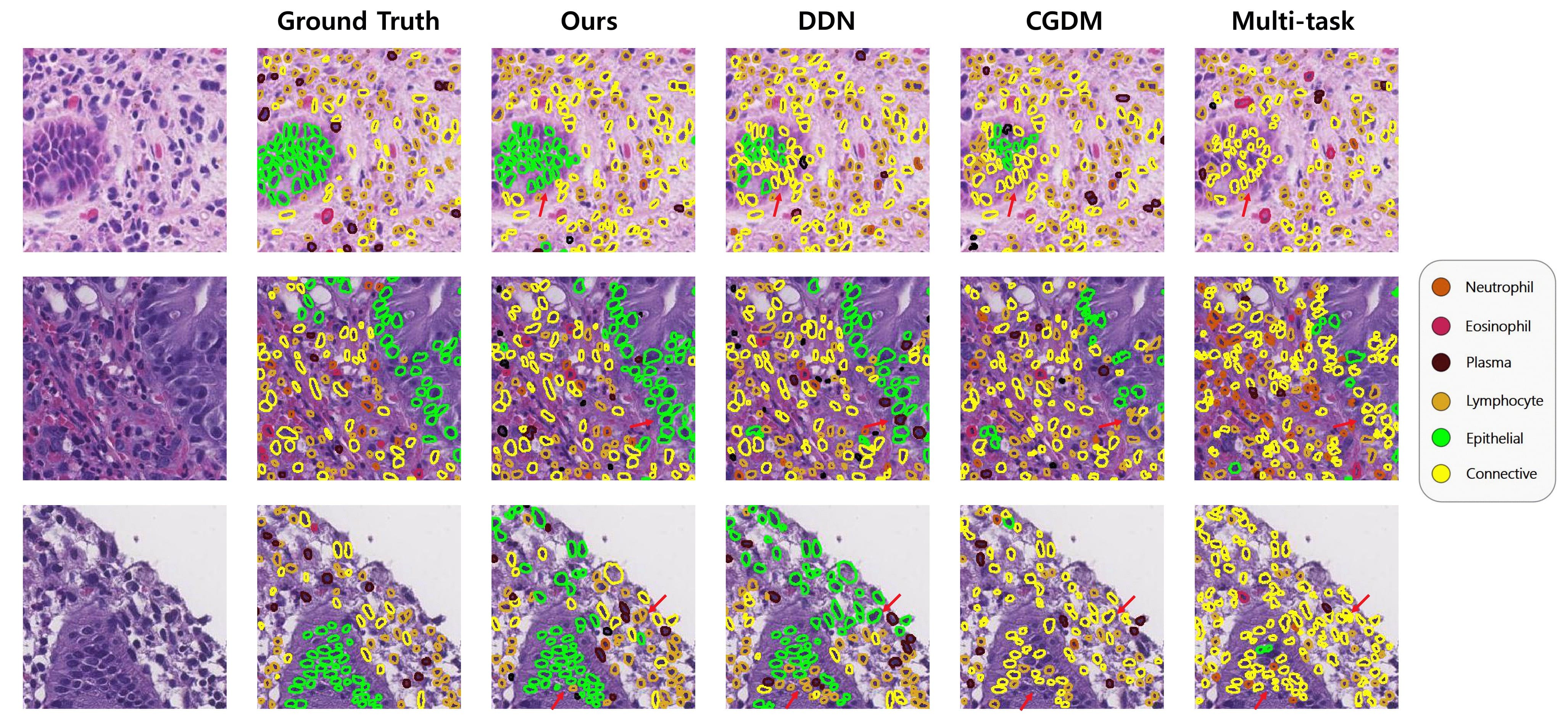}}
	\caption{Visual comparisons with other methods on the nuclei segmentation and recognition benchmark.}
	\label{fig:nuclei_result_vis}
\end{figure*}
\subsection{Skin Lesion Diagnosis}
\subsubsection{Datasets and Preprocessing}
Automatic fine-grained skin lesion recognition remains a global challenge in dermatology.
By taking a step further than the basic benign/malignant differentiation, identifying the specific subtypes of lesions demonstrates significant diagnostic value \cite{tschandl2020human}.
We hereby assign a benign/malignant discrimination dataset as the source domain, and a fine-grained multi-disease dataset as the target domain.
\textbf{HAM10000} is a dermatoscopic image dataset collected from different populations and modalities \cite{tschandl2018ham10000}.
After preprocessing procedures including histogram correction, sample filtering, and center crop, 10015 dermatoscopic images with lesions of seven diagnostic categories in total are provided.
It contains four subtypes of benign lesions\;(melanocytic nevi\;(NV), benign keratinocytic lesions\;(BKL), dermatofibromas\;(DF), and vascular lesions\;(VASC)) and three subtypes of malignant ones\;(melanomas\;(MEL), basal cell carcinomas\;(BCC), and actinic keratoses intraepithelial carcinomas\;(AKIEC)).
We use the face subset with only coarse two-class annotations as the source domain and the lower extremity subset with fine-grained seven-class annotations as the target domain.
All images are randomly cropped to the size of $160\times 160$ pixels before being forwarded to the network.

\subsection{Overall Experiment Settings}
For all experiments, we implement our method with Pytorch and conduct training on a NVIDIA GeForce RTX 3090 GPU with 24GB of memory.
Gradient backpropagation is performed for each mini-batch using BackPACK \cite{Dangel2020BackPACK}.
Following previous works in UDA \cite{chen2019synergistic}, each dataset is randomly split into $80\%/20\%$ as the training/test sets.
For novel classes in the target dataset, we sample few\;(5/10) samples with corresponding labels to formulate the support set.
The remaining target data is left unlabeled.
Data augmentation techniques such as rotation and horizontal/vertical flip are employed during training.
Please refer to the source code for more details.

It is noted that although from the technical perspective, skin lesion diagnosis is a multi-class classification problem similar to cancer tissue phenotyping, they differ largely in the task context.
Specifically, skin lesion diagnosis is more like an object recognition task where its decision is dominated by the local attributes of lesions,
while cancer tissue phenotyping relies on the global structure of the whole pathology images, instead of focusing on a salient object.

\begin{figure}[!t]
	\centerline{\includegraphics[width=\columnwidth]{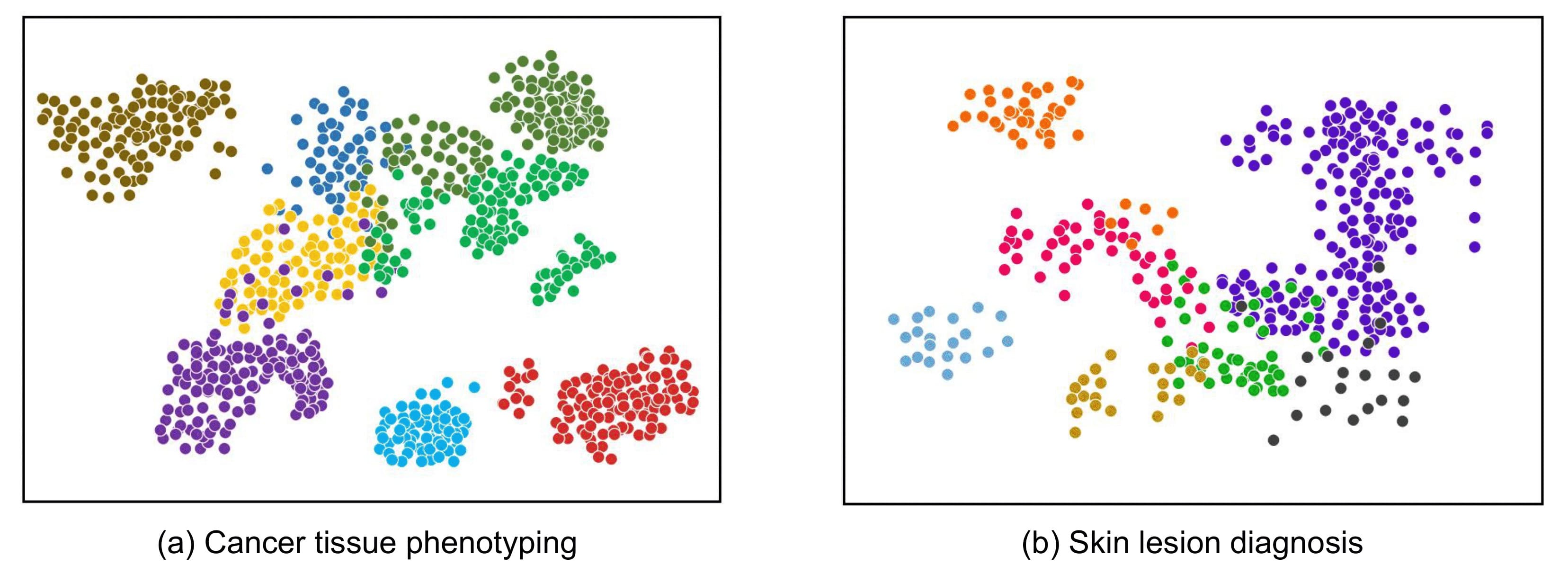}}
	\caption{Qualitative visualizations of our proposed method on the cancer tissue phenotyping and skin lesion diagnosis benchmark with t-SNE plot. Varied colours of points indicate the samples of different classes.}
	\label{fig:classification-tsne}
\end{figure}
\begin{table*}[!h]
	\centering
	\fontsize{8.5}{9.5}\selectfont
	\begin{threeparttable}
		\caption{Comparison results of our proposed method against other state-of-the-art methods for nuclei segmentation and recognition with 30-shot labeled target samples. The best and second-best results are highlighted in bold and brown, respectively.}
		\label{tab:nuclei_result_30}
		\setlength{\tabcolsep}{3mm}{
			\begin{tabular}{ccccc}
				\toprule[0.5mm]
				\multirow{3}{*}{\footnotesize \textbf{Methods}}
				&\multicolumn{4}{c}{\textbf{30-shot}}\cr
				\cmidrule(lr){2-5} 
				&mF1&mF1*&mPQ&mPQ*\cr
				\midrule[0.3mm]				
				Sup-only
				&43.78$_{0.56}$&36.73$_{0.34}$&21.83$_{0.54}$&21.27$_{0.38}$\cr
				Multi-task\,\cite{sener2018multi}
				&43.23$_{1.02}$&30.39$_{1.37}$&23.55$_{0.49}$&17.78$_{0.72}$\cr
				\midrule[0.2mm]			
				DANN\,\cite{ganin2016domain}
				&41.70$_{1.41}$&28.25$_{1.63}$&22.38$_{0.57}$&15.99$_{0.80}$\cr
				CGDM\,\cite{du2021cross}
				&43.82$_{0.98}$&34.36$_{0.91}$&24.92$_{0.34}$&20.48$_{0.26}$\cr
				\midrule[0.2mm]						
				LETR\,\cite{luo2017label}
				&40.04$_{1.13}$&25.80$_{0.82}$&21.96$_{0.70}$&14.89$_{0.44}$\cr
				FT-CIDA\,\cite{kundu2020class}
				&40.59$_{0.87}$&32.95$_{0.54}$&22.37$_{0.58}$&19.34$_{0.30}$\cr
				STARTUP\,\cite{phoo2021selftraining}
				&\textcolor{brown}{49.00$_{0.74}$}&36.32$_{1.10}$&\textcolor{brown}{28.78$_{0.46}$}&22.57$_{0.43}$\cr	
				DDN\,\cite{islam2021dynamic}
				&48.24$_{0.79}$&36.77$_{0.85}$&27.90$_{0.61}$&22.64$_{0.28}$\cr	
				TSA\,\cite{li2022crossfew}
				&44.96$_{0.60}$&33.13$_{0.97}$&25.69$_{0.91}$&21.40$_{0.63}$\cr	
				TACS\,\cite{gong2022tacs}
				&47.55$_{0.81}$&\textcolor{brown}{37.38$_{1.55}$}&28.47$_{0.83}$&\textcolor{brown}{23.18$_{1.04}$}\cr	
				\rowcolor{light-gray} Ours
				&\textbf{51.69}$_{0.48}$&\textbf{41.63}$_{0.65}$&\textbf{29.42}$_{0.55}$&\textbf{25.34}$_{0.67}$\cr	
				\bottomrule[0.5mm]
		\end{tabular}}
	\end{threeparttable}
\end{table*}

\begin{table}[!t]
	\centering
	\fontsize{8.5}{9.5}\selectfont
	\begin{threeparttable}
		\caption{\textcolor{black}{Comparison results of our proposed method against other state-of-the-art methods on three diverse tasks.}
		}
		\label{tab:extended_exp}
		\setlength{\tabcolsep}{1.7mm}{
			\begin{tabular}{ccccccc}
				\toprule[0.2mm]
				\multirow{3}{*}{\textbf{Methods}}
				&\multicolumn{2}{c}{\textbf{Radiology}}
				&\multicolumn{2}{c}{\textbf{Fundus}}
				&\multicolumn{2}{c}{\textbf{OfficeHome}}\cr
				\cmidrule(lr){2-3} \cmidrule(lr){4-5} \cmidrule(lr){6-7}
				\vspace{-0.8mm}
				&mF1&mF1*&mF1&mF1*&mF1&mF1*\cr
				\midrule[0.1mm]			
				Baseline&41.87&19.20&41.03&29.07&44.87&46.54\cr		
				CIDA\,[33]&42.55&23.48&39.90&27.32&43.29&42.80\cr
				TACS\,[19]&46.36&22.13&44.84&32.68&42.25&47.71\cr
				Ours&\textbf{49.23}&\textbf{26.54}&\textbf{46.26}&\textbf{37.71}&\textbf{50.08}& \textbf{54.67}
				\cr	
				\bottomrule[0.2mm]
		\end{tabular}}
	\end{threeparttable}
	\vspace{-4mm}
\end{table}

\begin{table*}[!t]
	\centering
	\fontsize{8.5}{9.5}\selectfont
	\begin{threeparttable}
		\caption{Ablation study by evaluating the performance gain of adding each key component.
			We start from the multi-task baseline \cite{sener2018multi} and add components cumulatively.
			Experiments are conducted on the nuclei segmentation and recognition benchmark.
		}
		\label{tab:ablation_add}
		\setlength{\tabcolsep}{2mm}{
			\begin{tabular}{ccccccccc}
				\toprule[0.5mm]
				\multirow{3}{*}{\footnotesize \textbf{Methods}}
				&\multicolumn{4}{c}{\textbf{5-shot}}
				&\multicolumn{4}{c}{\textbf{10-shot}}\cr
				\cmidrule(lr){2-5} \cmidrule(lr){6-9} 
				&mF1&mF1*&mPQ&mPQ*&mF1&mF1*&mPQ&mPQ*\cr
				\midrule[0.3mm]				
				Baseline
				&33.85&18.15&18.58&10.77
				&35.15&21.29&19.14&12.89\cr	
				+cross-domain
				&36.56&21.41&20.09&12.41
				&38.94&23.78&21.93&13.95\cr	
				+cross-class
				&36.40&22.84&20.15&13.30
				&40.52&27.26&22.78&16.13\cr	
				+historical
				&37.38&24.90&19.71&13.17
				&40.06&28.80&22.10&17.31\cr	
				+projection
				&\textbf{40.26}&\textbf{27.14}&\textbf{21.78}&\textbf{14.96}
				&\textbf{43.88}&\textbf{31.43}&\textbf{24.81}&\textbf{19.35}\cr	
				\bottomrule[0.5mm]
		\end{tabular}}
	\end{threeparttable}
\end{table*}

\section{Additional Experiment Results}
\subsection{Visualization}
We provide additional qualitative results on the three benchmarks.
The comparison results shown in Fig.\;\ref{fig:nuclei_result_vis} demonstrate the superiority of our method to detect each nucleus and delineate its boundary, as well as differentiating nuclei of various types with their detailed biological features.
The t-SNE visualization in Fig.\;\ref{fig:classification-tsne} shows that our method could discover the underlying embedding structures of various classes even with very limited labeled data.

\subsection{Results with More Annotations}
In this section, we compare our method with previous state-of-the-art approaches for cross-domain adaptation when more labeled samples are available in the target domain.
The results for nuclei segmentation and recognition with 30 labeled target samples are shown in Table\;\ref{tab:nuclei_result_30}.
The overall improvements of our method under this setting are consistent with previous experiment results.
It validates the effectiveness of our method under different levels of support.

\begin{figure*}[!t]
	\centerline{\includegraphics[width=1.5\columnwidth]{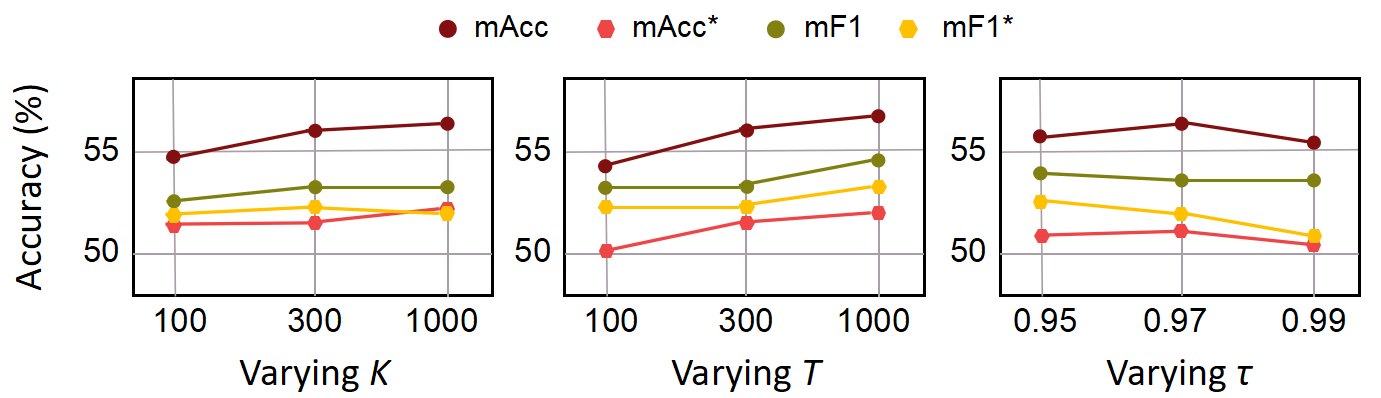}}
	\caption{Performance comparison between different choices of hyperparameters $K$, $T$ and $\tau$ on the cancer tissue phenotyping benchmark under 10-shot scheme.}
	\label{fig:further_parameter}
\end{figure*}
\begin{table*}[!t]
	\centering
	\fontsize{8.5}{9.5}\selectfont
	\begin{threeparttable}
		\caption{Comparisons of our proposed method against other state-of-the-art methods for nuclei segmentation and recognition by averaging the results from 10 random selections of support sets in the target domain. 
			The best results are highlighted in bold.
		}
		\label{tab:repeat_nuclei}
		\setlength{\tabcolsep}{1.3mm}{
			\begin{tabular}{ccccccccccccc}
				\toprule[0.5mm]
				\multirow{3}{*}{\footnotesize \textbf{Methods}}
				&\multicolumn{4}{c}{\textbf{5-shot}}
				&\multicolumn{4}{c}{\textbf{10-shot}}
				&\multicolumn{4}{c}{\textbf{30-shot}}\cr
				\cmidrule(lr){2-5} \cmidrule(lr){6-9} \cmidrule(lr){10-13} 
				&mF1&mF1*&mPQ&mPQ*&mF1&mF1*&mPQ&mPQ*&mF1&mF1*&mPQ&mPQ*\cr
				\midrule[0.3mm]				
				Multi-task\,\cite{sener2018multi}
				&30.04&16.17&16.51&9.20
				&33.58&20.73&18.45&12.40
				&39.52&28.16&22.21&17.25\cr	
				CGDM\,\cite{du2021cross}
				&33.30&19.88&19.08&12.46
				&37.63&26.69&20.61&15.21
				&42.53&35.11&23.96&20.82\cr	
				STARTUP\,\cite{phoo2021selftraining}
				&35.91&21.28&19.66&13.53
				&39.10&24.65&21.77&15.34
				&45.71&32.38&26.41&20.33\cr	
				Ours
				&\textbf{37.73}&\textbf{25.36}&\textbf{20.18}&\textbf{14.29}
				&\textbf{44.04}&\textbf{32.52}&\textbf{24.70}&\textbf{19.49}
				&\textbf{49.12}&\textbf{40.17}&\textbf{28.44}&\textbf{25.31}\cr
				\bottomrule[0.5mm]
		\end{tabular}}
	\end{threeparttable}
\end{table*}

\subsection{Extended Experiments on Diverse Tasks}
\textcolor{black}{We further evaluate our method on two medical image tasks beyond pathology analysis and one general visual recognition task, where the medical image tasks include pneumonia screening in radiology and diabetic retinopathy grading in fundus photography.
	For radiology analysis, we adopt covid-kaggle\,\cite{covid19} and Chest-Montreal\,\cite{cohen2020covid} for TADA from normal/pneumonia coarse screening to fine-grained pneumonia diagnosis.
	For fundus, DDR\,\cite{li2019diagnostic} and APTOS19\,\cite{aptos2019} are used to construct a TADA setting with two novel classes\;(grade level 3,\,4).
	In these settings, distribution shifts exist across domains due to differences in image acquisition protocols among multiple cohorts.
	For general visual task, we adopt OfficeHome\,\cite{venkateswara2017deep} and evaluate on ``Artist'' and ``Real-world'' domains.
	Experiments are conducted in 10-shot regime and the corresponding results are presented in Table\;\ref{tab:extended_exp}.
	Through comparison with SOTA methods, the effectiveness and broader applicability of our method are proved.}

\subsection{Extended Key Component Analysis}
In Table\;\ref{tab:ablation_add}, we demonstrate the effectiveness of our method's key components.
It complements the ablation study performed in the main text from a cumulative perspective.
From the results, we observe that all the proposed modules are beneficial for improving the cross-domain adaptation performance.
In particular, the employment of the cross-class distillation module contributes to a significant performance gain for target-private classes.
For instance, it attains $3.48\%$ and $2.18\%$ improvements in terms of mF1* and mPQ* under 10-shot scheme.
It verifies the effectiveness of our method to perform optimization trajectory distillation across domains and classes towards strong model generalization.

\subsection{Extended Hyperparameter Sensitivity Analysis}
We further analyze the choices of hyperparameters and their impacts on model performance in Fig.\;\ref{fig:further_parameter}.
We vary the values of $K$ and $T$, which denote the volumes of memory bank employed in the cross-domain/class distillation and historical self-distillation modules.
The choice of $\tau$ in Eq.\;(6) which coordinates the identification of principle subspace is also studied.
The results indicate that the choices of those hyperparameters do not have a significant influence as long as they are set within reasonable intervals.
Compared with them, the choices of $\lambda$ in Eq.\;(8) and $\kappa$ in Eq.\;(11) demonstrate more importance and are required to be carefully decided.

\subsection{Robustness to Support Sample Selection}
To evaluate the robustness of our method against the randomness during few-shot sample selection, we run the experiments for 10 times with different sets of labeled samples in the target domain.
The averaged results are presented in Table\;\ref{tab:repeat_nuclei}.
It demonstrates that our method consistently outperforms the competing approaches under diverse settings, which indicates its strong robustness.

{\small
	\bibliographystyle{ieee_fullname}
	\bibliography{egbib_MIA_submit,supplementary}

\begin{thebibliography}{10}\itemsep=-1pt

\bibitem{arora2017generalization}
Sanjeev Arora, Rong Ge, Yingyu Liang, Tengyu Ma, and Yi Zhang.
\newblock Generalization and equilibrium in generative adversarial nets (gans).
\newblock In {\em International Conference on Machine Learning}, pages
  224--232. PMLR, 2017.

\bibitem{azam2022recycling}
Sheikh~Shams Azam, Seyyedali Hosseinalipour, Qiang Qiu, and Christopher
  Brinton.
\newblock Recycling model updates in federated learning: Are gradient subspaces
  low-rank?
\newblock In {\em International Conference on Learning Representations}, 2022.

\bibitem{ben2006analysis}
Shai Ben-David, John Blitzer, Koby Crammer, and Fernando Pereira.
\newblock Analysis of representations for domain adaptation.
\newblock {\em Advances in neural information processing systems}, 19, 2006.

\bibitem{caldarola2022improving}
Debora Caldarola, Barbara Caputo, and Marco Ciccone.
\newblock Improving generalization in federated learning by seeking flat
  minima.
\newblock In {\em European Conference on Computer Vision}, pages 654--672.
  Springer, 2022.

\bibitem{cha2021swad}
Junbum Cha, Sanghyuk Chun, Kyungjae Lee, Han-Cheol Cho, Seunghyun Park, Yunsung
  Lee, and Sungrae Park.
\newblock Swad: Domain generalization by seeking flat minima.
\newblock {\em Advances in Neural Information Processing Systems},
  34:22405--22418, 2021.

\bibitem{chaudhari2017entropysgd}
Pratik Chaudhari, Anna Choromanska, Stefano Soatto, Yann LeCun, Carlo Baldassi,
  Christian Borgs, Jennifer Chayes, Levent Sagun, and Riccardo Zecchina.
\newblock Entropy-{SGD}: Biasing gradient descent into wide valleys.
\newblock In {\em International Conference on Learning Representations}, 2017.

\bibitem{chen2019synergistic}
Cheng Chen, Qi Dou, Hao Chen, Jing Qin, and Pheng-Ann Heng.
\newblock Synergistic image and feature adaptation: Towards cross-modality
  domain adaptation for medical image segmentation.
\newblock In {\em Proceedings of the AAAI conference on artificial
  intelligence}, volume~33, pages 865--872, 2019.

\bibitem{chen2022cross}
Wentao Chen, Zhang Zhang, Wei Wang, Liang Wang, Zilei Wang, and Tieniu Tan.
\newblock Cross-domain cross-set few-shot learning via learning compact and
  aligned representations.
\newblock In {\em European Conference on Computer Vision}, pages 383--399.
  Springer, 2022.

\bibitem{chen2018gradnorm}
Zhao Chen, Vijay Badrinarayanan, Chen-Yu Lee, and Andrew Rabinovich.
\newblock Gradnorm: Gradient normalization for adaptive loss balancing in deep
  multitask networks.
\newblock In {\em International conference on machine learning}, pages
  794--803. PMLR, 2018.

\bibitem{cheraghian2021synthesized}
Ali Cheraghian, Shafin Rahman, Sameera Ramasinghe, Pengfei Fang, Christian
  Simon, Lars Petersson, and Mehrtash Harandi.
\newblock Synthesized feature based few-shot class-incremental learning on a
  mixture of subspaces.
\newblock In {\em Proceedings of the IEEE/CVF International Conference on
  Computer Vision}, pages 8661--8670, 2021.

\bibitem{cohen2020covid}
Joseph~Paul Cohen.
\newblock Covid-19 image data collection: Prospective predictions are the
  future.
\newblock {\em arXiv:2006.11988}, 2020.

\bibitem{Dangel2020BackPACK}
Felix Dangel, Frederik Kunstner, and Philipp Hennig.
\newblock Backpack: Packing more into backprop.
\newblock In {\em International Conference on Learning Representations}, 2020.

\bibitem{das2022confess}
Debasmit Das, Sungrack Yun, and Fatih Porikli.
\newblock Confe{SS}: A framework for single source cross-domain few-shot
  learning.
\newblock In {\em International Conference on Learning Representations}, 2022.

\bibitem{dodd2018taxonomy}
Susanna Dodd, Mike Clarke, Lorne Becker, Chris Mavergames, Rebecca Fish, and
  Paula~R Williamson.
\newblock A taxonomy has been developed for outcomes in medical research to
  help improve knowledge discovery.
\newblock {\em Journal of clinical epidemiology}, 96:84--92, 2018.

\bibitem{dominguez2022cross}
C Dom{\'\i}nguez~Conde, C Xu, LB Jarvis, DB Rainbow, SB Wells, T Gomes, SK
  Howlett, O Suchanek, K Polanski, HW King, et~al.
\newblock Cross-tissue immune cell analysis reveals tissue-specific features in
  humans.
\newblock {\em Science}, 376(6594), 2022.

\bibitem{du2021cross}
Zhekai Du, Jingjing Li, Hongzu Su, Lei Zhu, and Ke Lu.
\newblock Cross-domain gradient discrepancy minimization for unsupervised
  domain adaptation.
\newblock In {\em Proceedings of the IEEE/CVF conference on computer vision and
  pattern recognition}, pages 3937--3946, 2021.

\bibitem{gamper2019pannuke}
Jevgenij Gamper, Navid Alemi~Koohbanani, Ksenija Benet, Ali Khuram, and Nasir
  Rajpoot.
\newblock Pannuke: an open pan-cancer histology dataset for nuclei instance
  segmentation and classification.
\newblock In {\em European congress on digital pathology}, pages 11--19.
  Springer, 2019.

\bibitem{ganin2016domain}
Yaroslav Ganin, Evgeniya Ustinova, Hana Ajakan, Pascal Germain, Hugo
  Larochelle, Fran{\c{c}}ois Laviolette, Mario Marchand, and Victor Lempitsky.
\newblock Domain-adversarial training of neural networks.
\newblock {\em The journal of machine learning research}, 17(1):2096--2030,
  2016.

\bibitem{gao2021gradient}
Zhiqiang Gao, Shufei Zhang, Kaizhu Huang, Qiufeng Wang, and Chaoliang Zhong.
\newblock Gradient distribution alignment certificates better adversarial
  domain adaptation.
\newblock In {\em Proceedings of the IEEE/CVF International Conference on
  Computer Vision}, pages 8937--8946, 2021.

\bibitem{Ge2020Mutual}
Yixiao Ge, Dapeng Chen, and Hongsheng Li.
\newblock Mutual mean-teaching: Pseudo label refinery for unsupervised domain
  adaptation on person re-identification.
\newblock In {\em International Conference on Learning Representations}, 2020.

\bibitem{gong2022tacs}
Rui Gong, Martin Danelljan, Dengxin Dai, Danda~Pani Paudel, Ajad Chhatkuli,
  Fisher Yu, and Luc Van~Gool.
\newblock Tacs: Taxonomy adaptive cross-domain semantic segmentation.
\newblock In {\em European Conference on Computer Vision}, pages 19--35.
  Springer, 2022.

\bibitem{graham2021lizard}
Simon Graham, Mostafa Jahanifar, Ayesha Azam, Mohammed Nimir, Yee-Wah Tsang,
  Katherine Dodd, Emily Hero, Harvir Sahota, Atisha Tank, Ksenija Benes, et~al.
\newblock Lizard: a large-scale dataset for colonic nuclear instance
  segmentation and classification.
\newblock In {\em Proceedings of the IEEE/CVF International Conference on
  Computer Vision}, pages 684--693, 2021.

\bibitem{graham2019hover}
Simon Graham, Quoc~Dang Vu, Shan E~Ahmed Raza, Ayesha Azam, Yee~Wah Tsang,
  Jin~Tae Kwak, and Nasir Rajpoot.
\newblock Hover-net: Simultaneous segmentation and classification of nuclei in
  multi-tissue histology images.
\newblock {\em Medical Image Analysis}, 58:101563, 2019.

\bibitem{guo2020broader}
Yunhui Guo, Noel~C Codella, Leonid Karlinsky, James~V Codella, John~R Smith,
  Kate Saenko, Tajana Rosing, and Rogerio Feris.
\newblock A broader study of cross-domain few-shot learning.
\newblock In {\em European conference on computer vision}, pages 124--141.
  Springer, 2020.

\bibitem{hochreiter1997flat}
Sepp Hochreiter and J{\"u}rgen Schmidhuber.
\newblock Flat minima.
\newblock {\em Neural computation}, 9(1):1--42, 1997.

\bibitem{hu2022adversarial}
Yanxu Hu and Andy~J Ma.
\newblock Adversarial feature augmentation for cross-domain few-shot
  classification.
\newblock In {\em European Conference on Computer Vision}, pages 20--37.
  Springer, 2022.

\bibitem{islam2021dynamic}
Ashraful Islam, Chun-Fu~Richard Chen, Rameswar Panda, Leonid Karlinsky, Rogerio
  Feris, and Richard~J Radke.
\newblock Dynamic distillation network for cross-domain few-shot recognition
  with unlabeled data.
\newblock {\em Advances in Neural Information Processing Systems},
  34:3584--3595, 2021.

\bibitem{izmailov2018averaging}
Pavel Izmailov, Dmitrii Podoprikhin, Timur Garipov, Dmitry Vetrov, and
  Andrew~Gordon Wilson.
\newblock Averaging weights leads to wider optima and better generalization.
\newblock {\em Proceedings of the Conference on Uncertainty in Artificial
  Intelligence}, 2018.

\bibitem{jacot2018neural}
Arthur Jacot, Franck Gabriel, and Cl{\'e}ment Hongler.
\newblock Neural tangent kernel: Convergence and generalization in neural
  networks.
\newblock {\em Advances in neural information processing systems}, 31, 2018.

\bibitem{jastrzkebski2018three}
Stanis{\l}aw Jastrz{\k{e}}bski, Zachary Kenton, Devansh Arpit, Nicolas Ballas,
  Asja Fischer, Yoshua Bengio, and Amos Storkey.
\newblock Three factors influencing minima in sgd.
\newblock {\em arXiv preprint arXiv:1711.04623}, 2017.

\bibitem{javed2020cellular}
Sajid Javed, Arif Mahmood, Muhammad~Moazam Fraz, Navid~Alemi Koohbanani,
  Ksenija Benes, Yee-Wah Tsang, Katherine Hewitt, David Epstein, David Snead,
  and Nasir Rajpoot.
\newblock Cellular community detection for tissue phenotyping in colorectal
  cancer histology images.
\newblock {\em Medical image analysis}, 63:101696, 2020.

\bibitem{aptos2019}
Maggie Karthik.
\newblock Aptos 2019 blindness detection.
\newblock {\em Kaggle}.

\bibitem{kather2016multi}
Jakob~Nikolas Kather, Cleo-Aron Weis, Francesco Bianconi, Susanne~M Melchers,
  Lothar~R Schad, Timo Gaiser, Alexander Marx, and Frank~Gerrit Z{\"o}llner.
\newblock Multi-class texture analysis in colorectal cancer histology.
\newblock {\em Scientific reports}, 6(1):1--11, 2016.

\bibitem{keskar2017on}
Nitish~Shirish Keskar, Dheevatsa Mudigere, Jorge Nocedal, Mikhail Smelyanskiy,
  and Ping Tak~Peter Tang.
\newblock On large-batch training for deep learning: Generalization gap and
  sharp minima.
\newblock In {\em International Conference on Learning Representations}, 2017.

\bibitem{kundu2020class}
Jogendra~Nath Kundu, Rahul~Mysore Venkatesh, Naveen Venkat, Ambareesh Revanur,
  and R~Venkatesh Babu.
\newblock Class-incremental domain adaptation.
\newblock In {\em European Conference on Computer Vision}, pages 53--69.
  Springer, 2020.

\bibitem{li2022domain}
Canran Li, Dongnan Liu, Haoran Li, Zheng Zhang, Guangming Lu, Xiaojun Chang,
  and Weidong Cai.
\newblock Domain adaptive nuclei instance segmentation and classification via
  category-aware feature alignment and pseudo-labelling.
\newblock In {\em International Conference on Medical Image Computing and
  Computer-Assisted Intervention}, pages 715--724. Springer, 2022.

\bibitem{li2019diagnostic}
Tao Li.
\newblock Diagnostic assessment of deep learning algorithms for diabetic
  retinopathy screening.
\newblock {\em Information Sciences}, 2019.

\bibitem{li2022crossfew}
Wei-Hong Li, Xialei Liu, and Hakan Bilen.
\newblock Cross-domain few-shot learning with task-specific adapters.
\newblock In {\em Proceedings of the IEEE/CVF Conference on Computer Vision and
  Pattern Recognition}, pages 7161--7170, 2022.

\bibitem{li2020hessian}
Xinyan Li, Qilong Gu, Yingxue Zhou, Tiancong Chen, and Arindam Banerjee.
\newblock Hessian based analysis of sgd for deep nets: Dynamics and
  generalization.
\newblock In {\em Proceedings of the 2020 SIAM International Conference on Data
  Mining}, pages 190--198. SIAM, 2020.

\bibitem{liang2021boosting}
Hanwen Liang, Qiong Zhang, Peng Dai, and Juwei Lu.
\newblock Boosting the generalization capability in cross-domain few-shot
  learning via noise-enhanced supervised autoencoder.
\newblock In {\em Proceedings of the IEEE/CVF International Conference on
  Computer Vision}, pages 9424--9434, 2021.

\bibitem{linder2012identification}
Nina Linder, Juho Konsti, Riku Turkki, Esa Rahtu, Mikael Lundin, Stig Nordling,
  Caj Haglund, Timo Ahonen, Matti Pietik{\"a}inen, and Johan Lundin.
\newblock Identification of tumor epithelium and stroma in tissue microarrays
  using texture analysis.
\newblock {\em Diagnostic pathology}, 7:1--11, 2012.

\bibitem{liu2022decompose}
Dongnan Liu, Chaoyi Zhang, Yang Song, Heng Huang, Chenyu Wang, Michael Barnett,
  and Weidong Cai.
\newblock Decompose to adapt: Cross-domain object detection via feature
  disentanglement.
\newblock {\em IEEE Transactions on Multimedia}, 25:1333--1344, 2022.

\bibitem{liu2020unsupervised}
Dongnan Liu, Donghao Zhang, Yang Song, Fan Zhang, Lauren O'Donnell, Heng Huang,
  Mei Chen, and Weidong Cai.
\newblock Unsupervised instance segmentation in microscopy images via panoptic
  domain adaptation and task re-weighting.
\newblock In {\em Proceedings of the IEEE/CVF conference on computer vision and
  pattern recognition}, pages 4243--4252, 2020.

\bibitem{luo2022channel}
Xu Luo, Jing Xu, and Zenglin Xu.
\newblock Channel importance matters in few-shot image classification.
\newblock In {\em International Conference on Machine Learning}, pages
  14542--14559. PMLR, 2022.

\bibitem{luo2017label}
Zelun Luo, Yuliang Zou, Judy Hoffman, and Li~F Fei-Fei.
\newblock Label efficient learning of transferable representations acrosss
  domains and tasks.
\newblock {\em Advances in neural information processing systems}, 30, 2017.

\bibitem{lupton2012medicine}
Deborah Lupton.
\newblock {\em Medicine as culture: illness, disease and the body}.
\newblock Sage, 2012.

\bibitem{mansilla2021domain}
Lucas Mansilla, Rodrigo Echeveste, Diego~H Milone, and Enzo Ferrante.
\newblock Domain generalization via gradient surgery.
\newblock In {\em Proceedings of the IEEE/CVF International Conference on
  Computer Vision}, pages 6630--6638, 2021.

\bibitem{milletari2016v}
Fausto Milletari, Nassir Navab, and Seyed-Ahmad Ahmadi.
\newblock V-net: Fully convolutional neural networks for volumetric medical
  image segmentation.
\newblock In {\em 2016 fourth international conference on 3D vision (3DV)},
  pages 565--571. Ieee, 2016.

\bibitem{panareda2017open}
Pau Panareda~Busto and Juergen Gall.
\newblock Open set domain adaptation.
\newblock In {\em Proceedings of the IEEE international conference on computer
  vision}, pages 754--763, 2017.

\bibitem{phoo2021selftraining}
Cheng~Perng Phoo and Bharath Hariharan.
\newblock Self-training for few-shot transfer across extreme task differences.
\newblock In {\em International Conference on Learning Representations}, 2021.

\bibitem{covid19}
Pranav Raikote.
\newblock Covid-19 image dataset.
\newblock {\em Kaggle}.

\bibitem{ravi2017optimization}
Sachin Ravi and Hugo Larochelle.
\newblock Optimization as a model for few-shot learning.
\newblock In {\em International Conference on Learning Representations}, 2017.

\bibitem{saito2019semi}
Kuniaki Saito, Donghyun Kim, Stan Sclaroff, Trevor Darrell, and Kate Saenko.
\newblock Semi-supervised domain adaptation via minimax entropy.
\newblock In {\em Proceedings of the IEEE/CVF International Conference on
  Computer Vision}, pages 8050--8058, 2019.

\bibitem{sener2018multi}
Ozan Sener and Vladlen Koltun.
\newblock Multi-task learning as multi-objective optimization.
\newblock {\em Advances in neural information processing systems}, 31, 2018.

\bibitem{shi2022gradient}
Yuge Shi, Jeffrey Seely, Philip Torr, Siddharth N, Awni Hannun, Nicolas
  Usunier, and Gabriel Synnaeve.
\newblock Gradient matching for domain generalization.
\newblock In {\em International Conference on Learning Representations}, 2022.

\bibitem{snell2017prototypical}
Jake Snell, Kevin Swersky, and Richard Zemel.
\newblock Prototypical networks for few-shot learning.
\newblock {\em Advances in neural information processing systems}, 30, 2017.

\bibitem{torralba2011unbiased}
Antonio Torralba and Alexei~A Efros.
\newblock Unbiased look at dataset bias.
\newblock In {\em CVPR 2011}, pages 1521--1528. IEEE, 2011.

\bibitem{tschandl2020human}
Philipp Tschandl, Christoph Rinner, Zoe Apalla, Giuseppe Argenziano, Noel
  Codella, Allan Halpern, Monika Janda, Aimilios Lallas, Caterina Longo, Josep
  Malvehy, et~al.
\newblock Human--computer collaboration for skin cancer recognition.
\newblock {\em Nature Medicine}, 26(8):1229--1234, 2020.

\bibitem{tschandl2018ham10000}
Philipp Tschandl, Cliff Rosendahl, and Harald Kittler.
\newblock The ham10000 dataset, a large collection of multi-source
  dermatoscopic images of common pigmented skin lesions.
\newblock {\em Scientific data}, 5(1):1--9, 2018.

\bibitem{Tseng2020Cross-Domain}
Hung-Yu Tseng, Hsin-Ying Lee, Jia-Bin Huang, and Ming-Hsuan Yang.
\newblock Cross-domain few-shot classification via learned feature-wise
  transformation.
\newblock In {\em International Conference on Learning Representations}, 2020.

\bibitem{venkateswara2017deep}
Hemanth Venkateswara.
\newblock Deep hashing network for unsupervised domain adaptation.
\newblock In {\em CVPR}, 2017.

\bibitem{wang2020unsupervised}
Qian Wang and Toby Breckon.
\newblock Unsupervised domain adaptation via structured prediction based
  selective pseudo-labeling.
\newblock In {\em Proceedings of the AAAI conference on artificial
  intelligence}, volume~34, pages 6243--6250, 2020.

\bibitem{wang2020generalizing}
Yaqing Wang, Quanming Yao, James~T Kwok, and Lionel~M Ni.
\newblock Generalizing from a few examples: A survey on few-shot learning.
\newblock {\em ACM computing surveys (csur)}, 53(3):1--34, 2020.

\bibitem{wang2019characterizing}
Zirui Wang, Zihang Dai, Barnab{\'a}s P{\'o}czos, and Jaime Carbonell.
\newblock Characterizing and avoiding negative transfer.
\newblock In {\em Proceedings of the IEEE/CVF conference on computer vision and
  pattern recognition}, pages 11293--11302, 2019.

\bibitem{winkler2019association}
Julia~K Winkler, Christine Fink, Ferdinand Toberer, Alexander Enk, Teresa
  Deinlein, Rainer Hofmann-Wellenhof, Luc Thomas, Aimilios Lallas, Andreas
  Blum, Wilhelm Stolz, et~al.
\newblock Association between surgical skin markings in dermoscopic images and
  diagnostic performance of a deep learning convolutional neural network for
  melanoma recognition.
\newblock {\em JAMA dermatology}, 155(10):1135--1141, 2019.

\bibitem{xu2021class}
Mengya Xu, Mobarakol Islam, Chwee~Ming Lim, and Hongliang Ren.
\newblock Class-incremental domain adaptation with smoothing and calibration
  for surgical report generation.
\newblock In {\em International Conference on Medical Image Computing and
  Computer-Assisted Intervention}, pages 269--278. Springer, 2021.

\bibitem{you2019universal}
Kaichao You, Mingsheng Long, Zhangjie Cao, Jianmin Wang, and Michael~I Jordan.
\newblock Universal domain adaptation.
\newblock In {\em Proceedings of the IEEE/CVF conference on computer vision and
  pattern recognition}, pages 2720--2729, 2019.

\bibitem{yu2020gradient}
Tianhe Yu, Saurabh Kumar, Abhishek Gupta, Sergey Levine, Karol Hausman, and
  Chelsea Finn.
\newblock Gradient surgery for multi-task learning.
\newblock {\em Advances in Neural Information Processing Systems},
  33:5824--5836, 2020.

\bibitem{yuan2022task}
Wang Yuan, Zhizhong Zhang, Cong Wang, Haichuan Song, Yuan Xie, and Lizhuang Ma.
\newblock Task-level self-supervision for cross-domain few-shot learning.
\newblock {\em Proceedings of the AAAI Conference on Artificial Intelligence},
  36(3):3215--3223, 2022.

\bibitem{yun2019cutmix}
Sangdoo Yun, Dongyoon Han, Seong~Joon Oh, Sanghyuk Chun, Junsuk Choe, and
  Youngjoon Yoo.
\newblock Cutmix: Regularization strategy to train strong classifiers with
  localizable features.
\newblock In {\em Proceedings of the IEEE/CVF international conference on
  computer vision}, pages 6023--6032, 2019.

\bibitem{zhang2019category}
Qiming Zhang, Jing Zhang, Wei Liu, and Dacheng Tao.
\newblock Category anchor-guided unsupervised domain adaptation for semantic
  segmentation.
\newblock {\em Advances in neural information processing systems}, 32, 2019.

\bibitem{zuo2021challenging}
Lin Zuo, Mengmeng Jing, Jingjing Li, Lei Zhu, Ke Lu, and Yang Yang.
\newblock Challenging tough samples in unsupervised domain adaptation.
\newblock {\em Pattern Recognition}, 110:107540, 2021.

\end{thebibliography}
}


\end{document}